\documentclass[fleqn,10pt]{wlscirep}
\usepackage[utf8]{inputenc}
\usepackage[T1]{fontenc}
\usepackage[super]{nth}
\usepackage{array}
\usepackage{graphicx}
\usepackage{multirow}
\usepackage{lineno}
\usepackage[font={small,it}]{caption}

\title{Deepfake Detection by Human Crowds, Machines, and Machine-informed Crowds}

\author[a,1]{Matthew Groh}
\author[a]{Ziv Epstein} 
\author[b]{Chaz Firestone}
\author[a]{Rosalind Picard}

\affil[a]{Massachusetts Institute of Technology Media Lab}
\affil[b]{Johns Hopkins University}

\begin{abstract}
The recent emergence of machine-manipulated media raises an important societal question: how can we know if a video that we watch is real or fake? In two online studies with 15,016 participants, we present authentic videos and deepfakes and ask participants to identify which is which. We compare the performance of ordinary human observers against the leading computer vision deepfake detection model and find them similarly accurate while making different kinds of mistakes. Together, participants with access to the model's prediction are more accurate than either alone, but inaccurate model predictions often decrease participants' accuracy. To probe the relative strengths and weaknesses of humans and machines as detectors of deepfakes, we examine human and machine performance across video-level features, and we evaluate the impact of pre-registered randomized interventions on deepfake detection. We find that manipulations designed to disrupt visual processing of faces hinder human participants' performance while mostly not affecting the model's performance, suggesting a role for specialized cognitive capacities in explaining human deepfake detection performance.

\end{abstract}

\begin{document}

\maketitle
\section*{Introduction}

How do we tell the difference between the genuine and the artificial? The emergence of deepfakes – videos that have been manipulated by neural network models to either swap one individual's face for another, or alter the individual's face to make them appear to say something they have not said  – presents challenges both for individuals and for society at large. Whereas a video of an individual performing an action or making a statement has long been one of the strongest pieces of evidence that the relevant event actually occurred, deepfakes undermine this gold standard, with potentially alarming consequences~\cite{lazer2018science, chesney2019deep, paris_deepfakes_2019, leibowicz2021deepfake}. 

How should we best meet this new challenge of evaluating the authenticity of a video? One approach is to build automated deepfake detection systems that analyze videos and attempt to classify their authenticity. Recent advances in training neural networks for computer vision reveal that algorithms are capable of surpassing the performance of human experts in some complex strategy games~\cite{silver_mastering_2016,silver_mastering_2017} and medical diagnoses~\cite{esteva_dermatologist-level_2017,mckinney_international_2020}, so we might expect algorithms to be similarly capable of outperforming people at deepfake detection. Indeed, such computational methods often surpass human performance in detecting physical implausibility cues~\cite{farid2010image}, such as geometric inconsistencies of shadows, reflections, and distortions of perspective images~\cite{nightingale2019can, nightingale2017can, kasra2018seeing}. Similarly, face recognition algorithms often outperform forensic examiners (who are significantly better than ordinary people) at identifying whether pairs of face images show the same or different people \cite{phillips2018face}. This focus on automating the analysis of visual content has advantages over certain methods from traditional digital media forensics, which often rely on image metadata \cite{farid2019fake} that are not available for many of today's most concerning deepfakes, which typically appear first on social media platforms stripped of such metadata~\cite{guess2020exposure,lyu2020deepfake}. Moreover, metadata from an individual's decision to share on social media may not be a reliable predictor of media's veracity because social media tends to focus people's attention on factors other than truth and accuracy~\cite{pennycook2021shifting, pennycook2021psychology}.

\begin{figure*}[ht]
    \centering
    \includegraphics[width=0.49\textwidth]{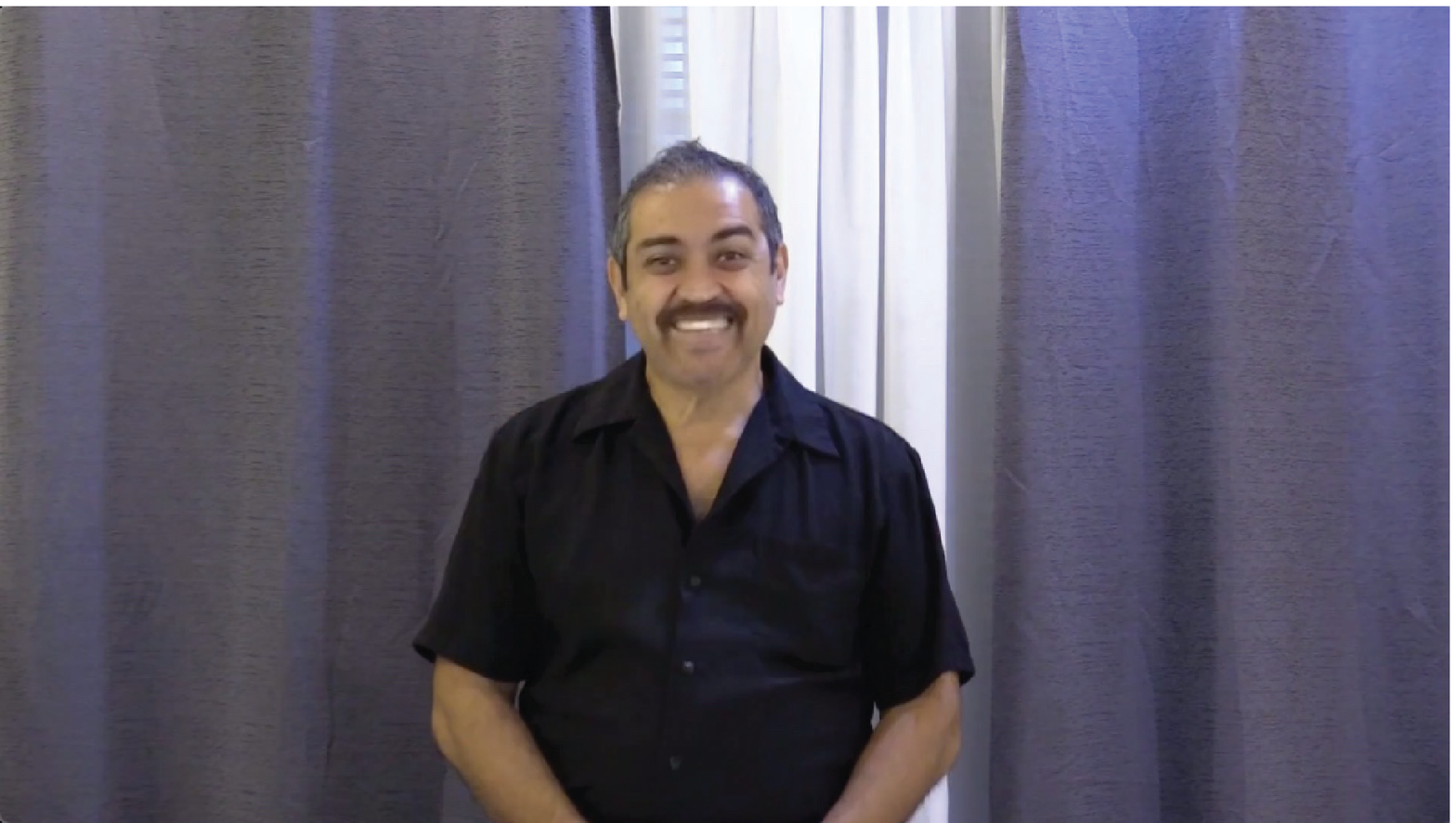}
    \includegraphics[width=0.49\textwidth]{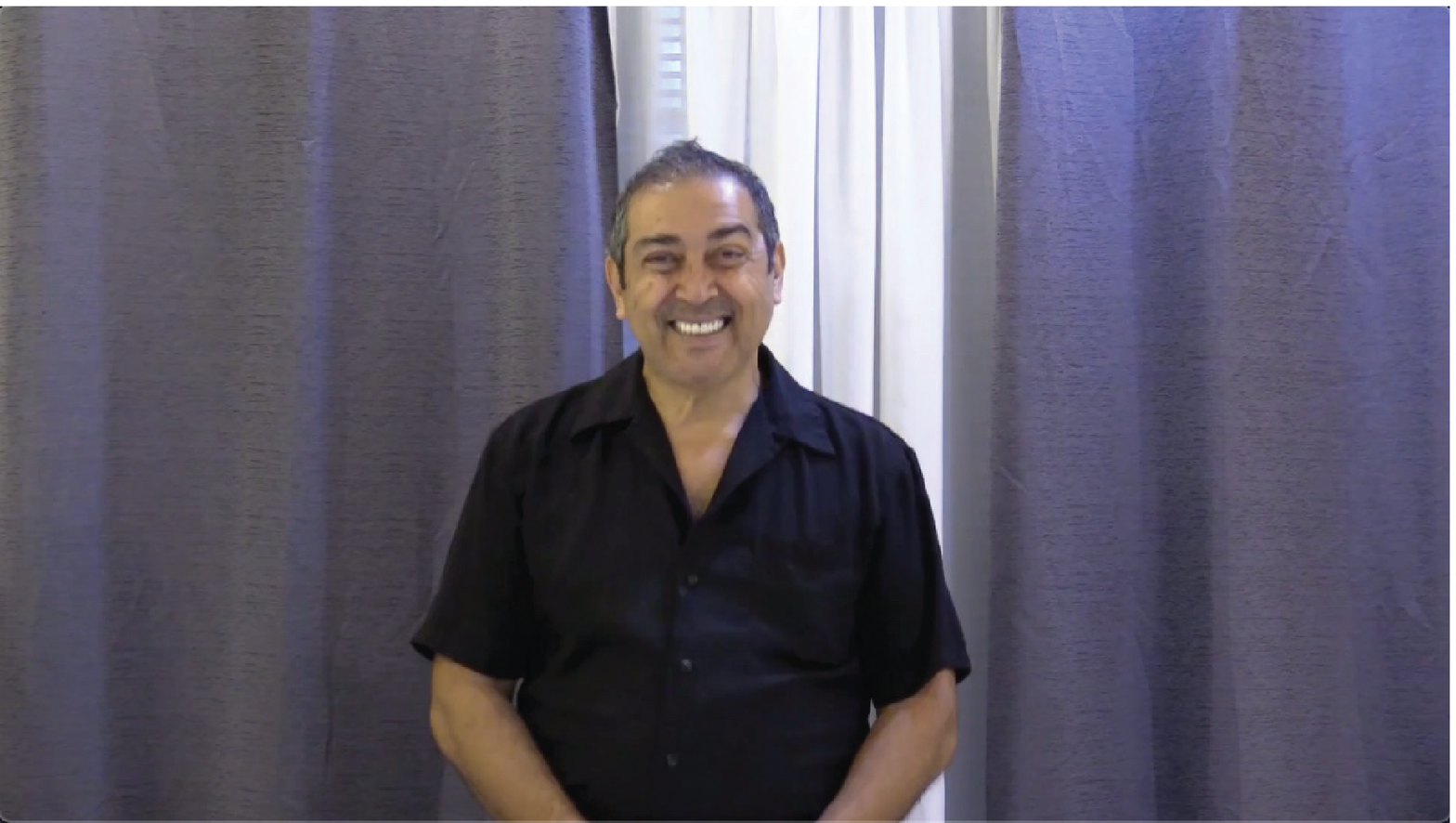}
    \caption{One of these two images is the first frame of a deepfake from Experiment 1; the other is the first frame of the original, authentic video from which the deepfake was created. Experiment 1 asked whether participants can tell which is which, using a two alternative forced-choice paradigm (i.e., selecting which of two video clips is a deepfake). Experiment 2 presented a single video and asked participants for their confidence the video is a deepfake or not. In this figure, the left panel is the deepfake; the man was not mustachioed at the time of filming.}
    \label{fig:deepfake_example}
\end{figure*}

The artificial intelligence (AI) approach to classifying videos as real or fake focuses on developing large datasets and training computer vision algorithms on these datasets~\cite{korshunov_deepfakes_2018, li_celeb-df_2020, yang_exposing_2018, rossler_faceforensics_2018, rossler2019faceforensics++, jiang_deeperforensics-10_2020, agarwal2020detecting, marra2018detection, mirsky2020creation, verdoliva2020media, tolosana2020deepfakes, dolhansky2019deepfake,dolhansky2020deepfake}. The largest open-source dataset is the Deepfake Detection Challenge (DFDC) dataset, which consists of 23,654 original videos showing 960 consenting individuals and 104,500 corresponding deepfake videos produced from the original videos. The first frames of both a deepfake and original video from this dataset appear in Figure~\ref{fig:deepfake_example}. The deepfakes examined here contain only visual manipulations produced using seven synthetic techniques: two deepfake autoencoders, a neural network face swap model~\cite{huang2012facial}, the NTH talking heads model~\cite{zakharov2019few}, the FSGAN method for reenactment and inpainting~\cite{nirkin2019fsgan}, StyleGAN~\cite{karras2019style}, and sharpening refinement on blended faces~\cite{dolhansky2020deepfake}. Unlike viral deepfake videos of politicians and other famous people, the videos from the competition have minimal context: they are all 10 second videos depicting unknown actors making uncontroversial statements in nondescript locations. As such, the cues for discerning real from fake can be based only on visual cues and not auditory cues or background knowledge of an individual or the topic they are discussing. In a contest run from 2019 to 2020, The Partnership for AI, in collaboration with large companies including Facebook, Microsoft, and Amazon, offered \$1,000,000 in prize money to the most accurate deepfake detection models on the DFDC holdout set via Kaggle, a machine learning competition website. A total of 2,116 teams submitted computer vision models to the competition, and the leading model achieved an accuracy score of 65\% on the 4,000 videos in the holdout data, which consisted of half deepfake and half real videos~\cite{dolhansky2020deepfake, dfdcresults2020}. While there are many proposed techniques for algorithmically detecting fakes (including affective computing approaches like examining heart rate and breathing rate~\cite{qi2020deeprhythm} and looking for emotion-congruent speech and facial expressions)~\cite{mittal2020emotions, agarwal_detecting_2020}, the most accurate computer vision model in the contest~\cite{selim2020} focused on locating faces in a sample of static frames using multitask cascaded convolutional neural networks~\cite{zhang2016joint}, conducting feature encoding based on EfficientNet B-7~\cite{tan2019efficientnet}, and training the model with a variety of transformations inspired by albumentations~\cite{buslaev_albumentations_2020} and grid-mask~\cite{chen_gridmask_2020}. Based on this model outperforming 2,115 other models to win significant prize money in a widely publicized competition on the largest dataset of deepfakes ever produced, we refer to this winning model as the ``leading model'' for detecting deepfakes to date. 

The rules of the competition strictly forbid human-in-the-loop approaches, which leaves open questions surrounding how well human-AI collaborative systems would perform at discerning between manipulated and authentic videos. In this paper, we address the following questions: How accurately do individuals detect deepfakes? Is there a ``wisdom of the crowds''~\cite{galton1907vox,surowiecki2005wisdom} effect when averaging participants' responses for each video? How does individual performance compare with the wisdom of the crowds, and how do these performances compare to the leading model's performance? Does access to the model's predictions and certainty levels help or hinder participants' discernment? And, what explains variation in human and machine performance; specifically, what is the role of video-level characteristics, can emotional priming influence participants' performance at detecting deepfakes, and does specialized processing of faces play a role in human and machine deepfake detection?

Crowdsourcing and averaging individuals' responses are promising and practical solutions for handling the scale of misinformation that would be otherwise overwhelming for an individual expert. Recent empirical research finds that averaged responses of ordinary people are on-par with third-party fact checkers for both factual claims in articles~\cite{allen2020scaling} and overall accuracy of content from URL domain names~\cite{epstein2020will, pennycook2019fighting}. In order to comprehensively compare humans to the leading AI model and evaluate collective intelligence against its artificial counterpart, we need to conduct two comparisons: How often do individuals outperform the model and how often does the the crowd wisdom outperform the model's prediction?

While a machine will consistently predict the same result for the same input, human judgment depends on a range of factors including emotion. Recent research in social psychology suggests that negative emotions can reduce gullibility~\cite{brashier2020judging, forgas2008being}, which could perhaps improve individual's discernment of videos. In particular, anger has been shown to reduce depth of thought by promoting reliance on stereotypes and previously held beliefs~\cite{clore2001affective}. Moreover, priming people with emotion has been demonstrated to both increase and decrease people's gullibility depending on the category of emotion~\cite{forgas2019happy} and hinder people's ability to discern real from fake news~\cite{martel2019reliance}. The role of emotion in deepfake detection is of practical concern because people share misinformation, especially political misinformation, because of its novelty and emotional content~\cite{vosoughi_spread_2018}. While a detailed examination of emotions as potential mechanisms to explain deepfake detection performance is outside the scope of this paper, we have included a pre-registered randomized experiment to evaluate whether experimentally elicited anger impairs participants' performance in detecting deepfakes. 

Based on research demonstrating human's expert visual processing of faces, we may expect humans to perform quite well at identifying the synthetic face manipulations in deepfake videos. Research in visual neuroscience and perceptual psychology has shown that the human visual system is equipped with mechanisms dedicated for face perception~\cite{sinha_face_2006}. For example, there is a region of the brain specialized for processing faces~\cite{kanwisher1997fusiform}. Human infants show sensitivity to faces even before being exposed to them~\cite{goren1975visual,reid2017human}, and adults are less accurate at recognizing faces when images are inverted or contain misaligned parts~\cite{yin1969looking,rhodes1993s,richler2011holistic}. The human visual system is faster and more efficient at locating human faces than other objects including objects with illusory faces~\cite{keys2021visual}. Whether human visual recognition of faces is an innate ability or a learned expertise through experience, visual processing of faces appears to proceed holistically for the vast majority of people ~\cite{richlerMetaanalysisReviewHolistic2014, young2018we}. In order to examine specialized processing of faces as a potential mechanism explaining deepfake detection performance, we include a randomized experiment where we obstruct specialized face processing by inverting, misaligning, and occluding videos.

In order to answer questions about human and machine performance at deepfake detection, we designed and developed a website called Detect Fakes where anyone on the internet could view deepfake videos sampled from the DFDC and see for themselves how difficult (or easy) it is to discern deepfakes from real videos. On this website, we conducted two randomized experiments to evaluate participants' ability to discern real videos from deepfakes and examine cognitive mechanisms explaining human and machine performance at detecting fake videos. We present a screenshot of the user interface of these two experiments in Figure S4 in the Supporting Information. In the first experiment, we present a two-alternative forced choice design where a deepfake video is presented alongside its corresponding real video. In the second experiment, we presented participants with a single video design and asked them to share how confident (from 50\% to 100\% in 1 percentage point increments) they are that the video is a deepfake (or is not a deepfake). In this single video framework, we present participants with the option to update their confidence after seeing the model's predicted likelihood that a video is a deepfake. By doing so, we evaluate how machine predictions affect human decision-making. In both experiments, we embedded randomized interventions to evaluate whether incidental emotion (emotion unrelated to the task at hand) or obstruction of specialized processing of faces influence participants' performance.

\section*{Results}

\subsection*{Experiment 1: Two-Alternative Forced Choice (N=5,524)}

In Experiment 1, 5,524 individuals found our website organically and participated in 26,820 trials. The 56 pairs of videos in Experiment 1 were sampled from the DFDC training dataset because the experiment was conducted before the holdout videos for the DFDC dataset were publicly released. As such, we compare participants' performance in Experiment 1 to the overall performance of the leading model. We leave a direct comparison of participant and model performance for Experiment 2 which focuses on performance across holdout videos. 

\subsubsection*{Individual vs. Machine}

\begin{figure*}[ht]
    \centering

    \includegraphics[width=0.98\textwidth]{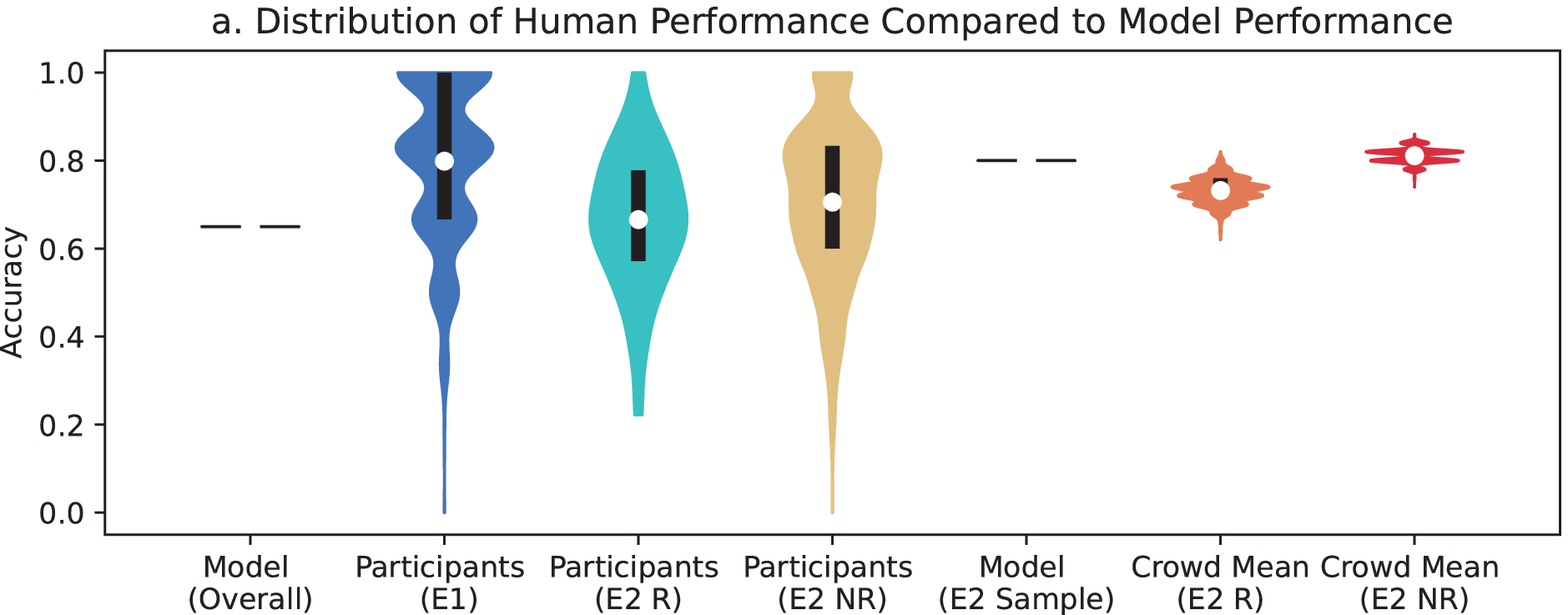}
    \includegraphics[width=0.32\textwidth]{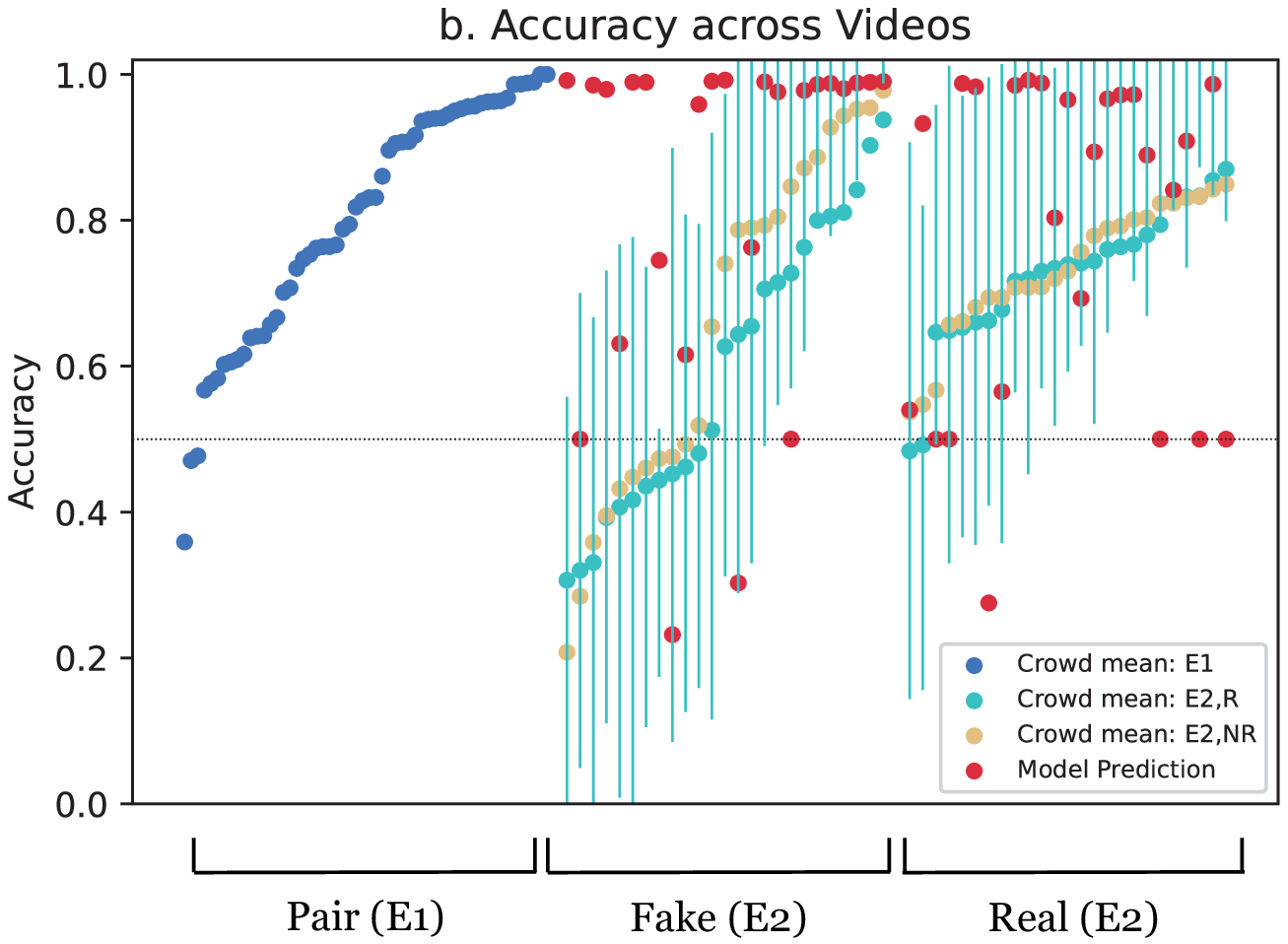}
    \includegraphics[width=0.32\textwidth]{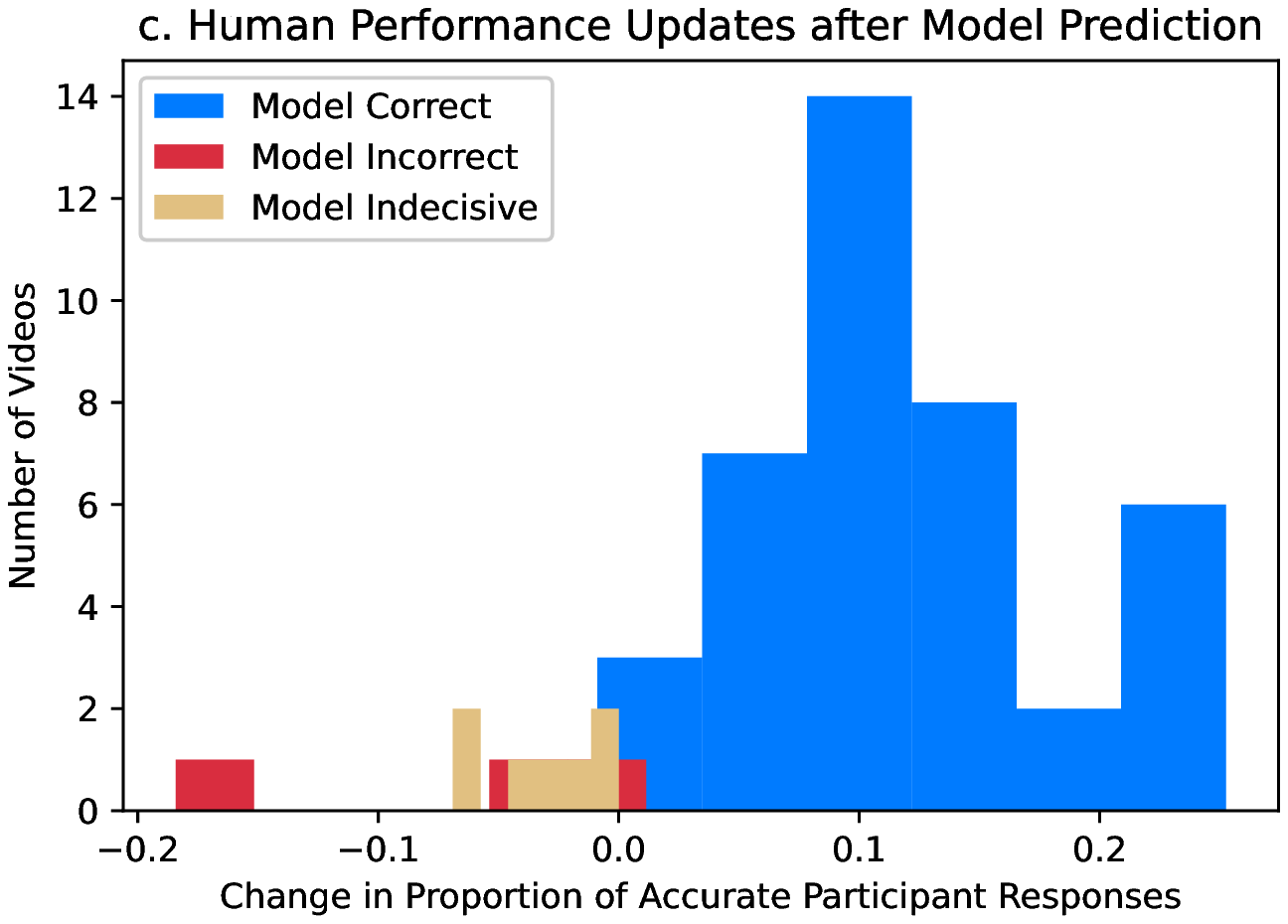}
    \includegraphics[width=0.32\textwidth]{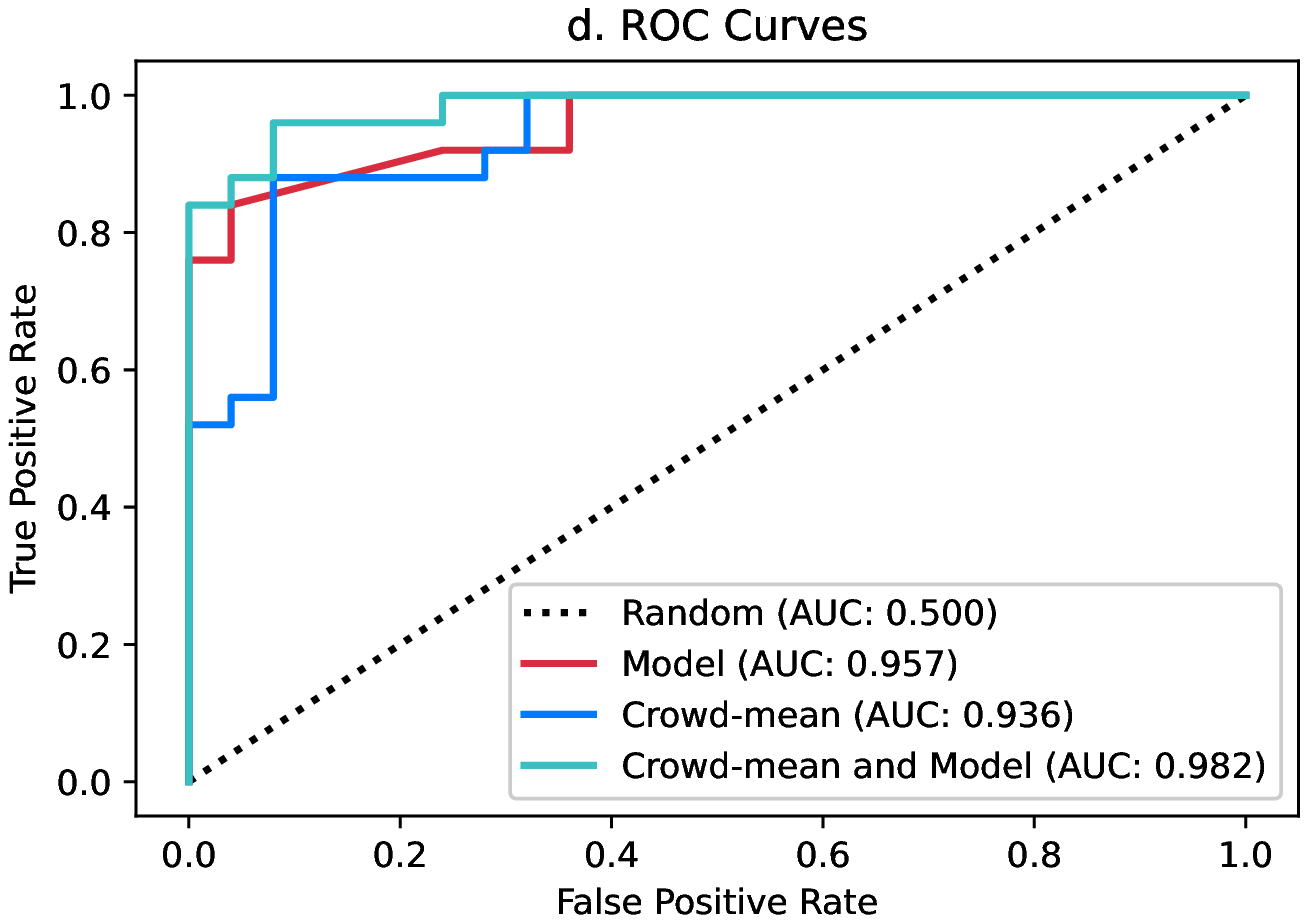}

    \caption{Figure 2a presents the distribution of participant performance across experiments compared to the model's performance via violin plots where the white dots indicate the mean and the black bars indicate the interquartile range. R refers to recruited participants, NR refers to non-recruited participants, E1 refers to Experiment 1, and E2 refers to Experiment 2. In the Experiment 1 (two-alternative forced choice), accuracy is defined as identifying a deepfake from a pair of videos correctly. In Experiment 2 (single video design), accurate identification is defined as responding with the correct answer with more than 50\% confidence. The model's performance represents a single observation in each instance, and as such, we present the model's performance as a horizontal black line with a white dot in the middle. The crowd mean distributions are obtained by bootstrapping confidence intervals based on 1000 randomly drawn samples that are each half of the total observations. Figure 2b presents a scatter plot of the model's accuracy and the distribution of participants' accuracy scores for each video. The x-axis of Figure 2b is an index of the videos, and it is ordered by experiment, true class of each video, and participant's average accuracy. The teal lines in Figure 2b represent the interquartile range of recruited participants' responses. Figure 2c presents the distribution of changes in recruited participants' accuracy after updating their response based on whether the model's prediction is correct, incorrect, or indecisive. Figure 2d presents the receiver operator characteristic curves of computer performance, recruited participants' collective performance, and recruited participants' collective performance with the model's decision support across the 50 DFDC videos in Experiment 2.}
    \label{fig:4graphs}
\end{figure*}

As stated in our pre-analysis plan\footnote{Pre-analysis plan for non-recruited participants in Experiment 1: https://aspredicted.org/blind.php?x=wg84ic} for Experiment 1, we examined the accuracy of all participants who saw at least 10 pairs of videos, for a total of 882 participants. 82\% of participants outperform the leading model, which achieves 65\% accuracy on the holdout dataset~\cite{dfdcresults2020}. Half of the stimuli set (28 of 56 pairs of videos) was identified correctly by over 83\% of participants, 16 pairs of videos were identified correctly by between 65\% and 83\% of participants, and 12 pairs of videos were identified correctly by less than 65\% of participants. Out of these 12 pairs of videos, 3 pairs of videos were identified correctly by less than 50\% of participants. Figure~\ref{fig:4graphs}a presents the distribution of participants' performance in Experiment 1 (in blue in the second column) next to the model's overall performance (in black in the first column). 

We do not find any evidence that participants improve in their ability to detect these videos within the first ten videos seen ($p=0.112$) (all p-values reported in this paper are generated by linear regression with robust standard errors clustered on participants unless otherwise stated). On average, participants took 42 seconds to respond to each pair of videos, and we find that for every additional ten seconds participants take to respond, participants' accuracy decreased by 1.1 percentage points  ($p<0.001$). We embedded three randomized experiments in Experiment 1 to evaluate the roles of specialized processing of faces, time for reflection, and emotion elicitation. We find participants are 5.6 percentage points less accurate ($p=0.004$) at detecting pairs of inverted videos than pairs of upright videos. In contrast, we do not find statistically significant effects of the additional time for reflection intervention or this particular emotion elicitation intervention. The custom emotion elicitation intervention in this first experiment did not have a statistically significant influence on participants' self-reported emotions, which suggests the custom emotion elicitation experiment did not work here. We provide additional details on the interventions in Experiment 1 in the Supplementary Information section.

\subsection*{Experiment 2 – Single Video Design (N=9,492)}

In Experiment 2, 9,492 individuals participated: 304 individuals were recruited from Prolific and completed 6,390 trials; 9,188 individuals found our website organically and completed 67,647 trials.\footnote{Pre-analysis plan for recruited individuals participating in Experiment 2: https://aspredicted.org/blind.php?x=mp6yg9} In the recruited cohort, all but 3 participants viewed 20 videos. In the non-recruited cohort, over half of participants viewed 7 videos and the \nth{90} percentile participant viewed 17 videos. The website instructed participants about videos that ``Half are deepfakes, half are not.'' After viewing each video, participants move a slider to report their response ranging from ``100\% confidence this is NOT a DeepFake'' to ``100\% confidence this is a DeepFake'' in one percent increments with ``just as likely a DeepFake as not'' in the middle (see Figure S4 in the Supporting Information for a screenshot of the user interface). Participants can never make a selection with less than 50\% confidence; the slider's default position is in the center (at the ``just as likely a DeepFake as not'' position and one increment to the right becomes ``51\% confidence this is a DeepFake'' and one increment to the left becomes ``51\% confidence this is NOT a DeepFake.'' The stimuli in Experiment 2 consist of 50 videos randomly sampled from the competition holdout dataset (half deepfake and half non-manipulated), 4 videos of Kim Jung-un and Vladimir Putin including both one deepfake and non-manipulated video of each leader, and a deepfake attention check video. 

In Experiment 2, we define the accuracy score as the participant's response between 0 and 1 normalized for correctness, which is the participant's response if correct or 1 minus the participant's response if incorrect. For example, if a participant responded ``82\% confidence this is a DeepFake'' and the participant is correct, then the participant is assigned an accuracy score of 0.82. If the participant is incorrect, then the participant would be assigned an accuracy score of 0.18. We define accurate identification as an accuracy score greater than 0.5.

Participants' and the leading model's performance on deepfake detection depends on the population of participants, the population of videos, how performance is measured at the individual or collective level, and whether videos are presented side by side or by themselves. In some cases, we find a machine advantage and in others, we find a human advantage. The rest of the results section examines individual participant performance compared with the leading model, participants' collective performance compared with the leading model, participants' collective performance when participants have access to the model's predictions, variations in human and machine performance across videos, and randomized experiments designed to evaluate the role of emotional priming and specialized visual processing of faces.

\subsubsection*{Individual vs. Machine}

For participants who pass the attention check, recruited participants accurately identified deepfakes from the randomly sampled holdout videos in 66\% of attempts while the non-recruited participants accurately identified videos in 69\% of trials (or 72\% of attempts when limiting the analysis to non-recruited participants who saw at least 10 videos). In comparison, the leading model accurately identified deepfakes on 80\% of the sampled videos, which is significantly better than the 65\% accuracy rate this model achieves on the full holdout dataset of 4,000 videos~\cite{dfdcresults2020}.

In a direct comparison of performance, 13\% of recruited participants, 27\% of non-recruited participants who saw at least 10 videos, and 37\% of non-recruited participants who saw fewer than 10 videos outperform the model. Figure~\ref{fig:4graphs}a presents the distribution of participants' accuracy on the sampled holdout videos (in teal for recruited participants and gold for non-recruited participants). Relative to the leading model, participants are less accurate at identifying deepfakes than they are at identifying real videos. Recruited participants accurately identify deepfakes as deepfakes in 57\% of attempts compared to the leading model identifying deepfakes as deepfakes in 84\% of videos while both recruited participants and the leading model identify real videos as real videos at nearly same rate (75\% of participants' observations and 76\% of videos). Recruited participants predicted the sampled holdout videos were real (57\% of observation) considerably more often than fake (38\% of observations) while the computer vision model predicted videos were real (44\% of observations) barely more frequently than fake (42\% of observations). In 5\% of recruited participant observations and 14\% of computer vision model observations, the prediction was a 50-50 split between real and fake. We report confusion matrices for each treatment condition in Tables 3-7 in the Supplemental Information.

On the additional set of videos of political leaders, participants outperform the leading model. Specifically, 60\% of recruited participants and 68\% of non-recruited participants who saw at least ten videos outperform the model on these videos. For the deepfake videos of Kim Jong-un, Vladimir Putin, and the attention check, the state-of-the-art computer vision model outputs a 2\%, 8\%, and 1\% probability score that each respective video is a deepfake, which is both confident and inaccurate.

We do not find any evidence that participants' overall accuracy changes as participants view more videos ($p=0.433$). However, we find that for every additional video seen by recruited participants, they are 0.9\% ($p<0.001$) more likely to report any video as a deepfake. This corresponds to performing about 18\% better at detecting deepfakes and 18\% worse at identifying real videos by the last video.

Recruited participants spent a median duration of 22 seconds before submitting their initial guess and a median duration of 3 seconds adjusting (or not adjusting) their initial guess when prompted with the model's predicted likelihood. Non-recruited participants spend a similar amount of time. For both sets of participants, we find that for every ten additional seconds of participant response time, participants' accuracy decreases by 1 percentage point ($p<0.001$).

\subsubsection*{Crowd Wisdom vs. Machine}

The crowd mean, participants' responses averaged per video, is on par with the leading model performance on the sampled holdout videos. For recruited participants, the crowd mean accurately identifies 76\% of videos. For non-recruited participants, the crowd mean accurately identifies 80\% of videos. For the 1,879 non-recruited participants who saw at least 10 videos, the crowd mean is 86\% accurate. In comparison, the leading model accurately identifies 80\% of videos. 

In Figure~\ref{fig:4graphs}b, we compare statistics on participants' accuracy (the mean and interquartile range) with the model's predictions for each video. In Table 2 in the Appendix, we present the mean accuracy of recruited and non-recruited participants and the computer vision model for all videos. There are 2 videos (both deepfakes) on which both the crowd mean and the leading model are at or below the 50\% threshold. One of these videos (video 7837) is extremely blurry, while the other video (video 4555) is filmed from a low angle and the actress's glasses show significant glare. 

There are 8 videos on which the crowd mean is accurate but the model is at or below the 50\% threshold and another 5-12 videos on which the model is accurate but the crowd mean (depending on how the population selected) is below the 50\% threshold.

\subsubsection*{Human-AI Collaboration}

In addition to comparing individual and collective performance to the leading model's performance, we examined how an AI model could complement human performance. After participants' submitted their initial response for how confident they are that a video is or is not a deepfake in Experiment 2, we revealed the likelihood that the video is a deepfake – as predicted by the leading model – and gave participants a chance to update their response. After taking into account the model's prediction, participants updated their confidence in 24\% of trials (and crossing the 50\% threshold for accurate identification in 12\% of trials). By updating their responses, recruited participant's accurate identification increased from 66\% to 73\% of observations ($p<0.001$ based on a Student's t-test). Figure~\ref{fig:4graphs}c presents the distribution of changes in overall participant accuracy for the 50 videos sampled from the DFDC. For the 40 videos upon which the model accurately identifies the video as a deepfake or not, participants updated their responses to be on average 10.4\% more accurate at identification than before seeing the model's prediction. For the remaining 10 videos on which the model made an incorrect or equivocal prediction, participants updated their responses to be on average 2.7\% less accurate at identification than before seeing the model's predictions. In the most extreme example of incorrect updating, the model predicted a 28\% likelihood the video was a deepfake when it was indeed a deepfake and participants updated their responses to be on average 18\% less accurate at identifying the deepfake. This particular video (video 7837) is quite blurry, and perhaps, participants changed their responses because it's very difficult to discern manipulations in low quality video. 

For the additional deepfake videos of Kim Jung-un and Vladimir Putin that are not included in the overall analysis, the model predicted a 2\% and 8\% likelihood the video was a deepfake, respectively. This prediction is not only incorrect but confidently so, which led participants to update their responses such that participant's accurate identification dropped from 56\% to 34\% on the Kim Jung-un deepfake and 70\% to 55\% on the Vladimir Putin deepfake. 

In Figure~\ref{fig:4graphs}d, the receiver operating characteristic (ROC) curve of the leading model is plotted alongside the ROC curves of the crowd mean and crowd-mean responses where participants have access to the model's prediction for each video. While the model has a slightly higher AUC score of 0.957 relative to the crowd means's AUC score of 0.936, either the model or the crowd mean could be considered to perform better depending on the acceptable false positive rate. However, the crowd mean response after seeing the model's predictions strictly outperforms both the performance of the crowd mean and leading model. When we condition the ROC analysis on confidence following methods for estimating the reliability of eyewitness identifications~\cite{wixted2016estimating}, we find that medium and high confidence responses outperform low confidence responses by a large degree. We define low confidence as responses between 33.5 and 66.5, medium confidence as responses between 17 and 33.5 or 66.5 and 83, and high confidence as responses between 0 and 17 or 83 and 100. Figure S2 in the Supporting Information ROC curves presents a visual comparison of model performance to low, medium, and high confidence responses from participants, which reveals medium and high (but not low) confidence responses can outperform the model's predictions depending on the acceptable false positive rate. 

\subsubsection*{Video Features Correlated with Accuracy}

Given the heterogeneity in both participants' and the leading model's performance on videos, we extend the analysis of performance across seven video-level features: graininess, blurriness, darkness, presence of a flickering face, presence of two people, presence of a floating distraction, and the presence of an individual with dark skin. These video-level characteristics were hand-labeled by the research team. On the 14 videos that are either grainy, blurry, or dark, the crowd wisdom of recruited participants is correct on 8 videos while the crowd wisdom of non-recruited participants and the model is correct on 10 videos. When we examine the 36 videos that are neither grainy, blurry, nor extremely dark, the crowd wisdom of recruited participants is correct on 29 out of 36 videos, the crowd wisdom of non-recruited participants is correct on 32 out of 36 videos, and the model is correct on 30 of 36 videos. The presence of a flickering face is associated with an increase in recruited participants' accuracy rates by 24.2 percentage points ($p<0.001$) and an increase in the model's accuracy rates by 16.9 percentage points ($p=0.007$) in detecting a deepfake. The presence of two people in a video instead of a single person is associated with an overall increase in recruited participants' accuracy rates by 7.6\% ($p<0.001$) and a 21.8\% decrease ($p=0.024$) by the model in identifying real videos. The presence of a floating distraction is associated with a decrease in recruited participants' accuracy rates on real videos of 3.5\% ($p=0.034$) and an increase in recruited participants' accuracy rates on fake videos of 11.3\% ($p<0.001$). In 12 of 50 videos, at least one person in the video has dark skin (precisely defined as skin classified as type 5 or 6 on the Fitzpatrick scoring system, which is a classification system developed for dermatology that computer vision researchers have used to examine the context of skin color)~\cite{buolamwini2018gender}. We find that the presence of an individual with dark skin in the video is associated with a decrease in recruited participants' accuracy by 8.8\% ($p<0.001$) and a decrease in the model's accuracy by 11.8\% ($p=0.187$). In order to control for these seven comparisons conducted simultaneously, we can apply a Bonferroni correction of 1/7 to the standard statistical significance thresholds (e.g., a p-value threshold of 0.01 becomes 0.0014). Based on this correction, the influence of a flickering face, two people in the same video, floating distractions, and the presence of an individual with dark skin continue to be statistically significant for participants if the original p-value threshold is chosen as 0.01.

\subsubsection*{Randomized Experiments for Evaluating Emotion Priming and Specialized Face Processing}

Within Experiment 2, we embedded two randomized experiments to examine potential cognitive mechanisms underpinning how humans discern between real and fake videos. Specifically, we examine an affective intervention designed to elicit anger based on a well-established intervention~\cite{small2008emotional} (see Figure S3 in the Supporting Information) and a perceptual intervention designed to obstruct specialized processing of faces via inversion (videos presented upside down), misalignment (videos presented with actors' faces horizontally split), and occlusion (videos presented with a black bar over the actors' eyes).

We present results of the anger elicitation intervention in columns 1-3 in Table~\ref{fig:reg1}. We do not find statistically significant effects ($p=0.280$) of the anger elicitation intervention on overall accuracy. However, in our pre-registered follow-up analysis limiting the dataset to real videos, we find that participants who were assigned to the anger elicitation treatment underperformed control participants by 5.2 percentage points ($p=0.032$). In other words, participants in the anger elicitation treatment are more 5.2 percentage points more likely than participants in the control group to make a false positive identification that a real video is a deepfake. Notice here that the floor is not 0\% accuracy but rather 50\% accuracy (i.e., chance responding); the maximum effect of the anger elicitation treatment is 21.6 percentage points (71.6 from the constant term in column 2 of Table~\ref{fig:reg1} minus 50), so a 5.2 percentage point reduction represents an effect that is 24.1\% of the maximum possible effects under these conditions.

Figure S3 in the Supporting Information presents accuracy and confidence scores by treatment assignment to visually reveal the heterogeneous effect of anger elicitation on how participants discern between real and fake videos. When we examine the relationship between assignment to the anger elicitation group and how confident participants guess, we do not find a statistically significant effect ($p=0.347$). When we examine real videos and the relationship between anger elicitation and updating predictions after seeing what the model would predict, we find that participants assigned to the anger elicitation group are 3.7\% ($p=0.035$) more likely to change their guess to a correct answer than participants assigned to the control group. As a result, we do not find statistically significant effects of anger elicitation on accuracy after participants update their response ($p=0.246$).

We present results of the perceptual obstruction intervention in columns 1-6 in Table~\ref{fig:reg1}. We find statistically significant effects of all three specialized processing obstructions on participants' ability to accurately identify deepfakes from authentic videos. The overall effects – reported in column 1 of Table~\ref{fig:reg1} – are all statistically significant and range from a decrease of 4.3 percentage points in accuracy for the inversion treatment ($p=0.002$) to a decrease of 4.4 percentage points in accuracy for the eye occlusion treatment ($p=0.004$) to a decrease of 6.3 percentage points for the misalignment treatment ($p<0.001$) on a base rate of 65.5\% accuracy when controlling for video fixed effects. In addition, we find that inverting the videos decreases participants' reported confidence scores (absolute distance in guesses from the 50-50 selection) by 2 percentage points ($p=0.002$), but we do not find similar decreases in reported confidence on videos with misalignment or occlusion transformations. 

In the sample of recruited participants, the specialized face processing obstructions have different effects depending on whether the videos are manipulated or not. When we limit the analysis to the algorithmically manipulated deepfakes (column 3 of Table~\ref{fig:reg1}), we do not find statistically significant effects on the inversion treatment ($p=0.638$) but we do find that the misalignment and eye occlusion treatments show a decrease in 7.7 ($p=0.002$) and 6.3 ($p=0.008$) percentage points, respectively, relative to the control videos. In contrast, when we limit the analysis to the other half of videos that have not been manipulated (column 2 of Table~\ref{fig:reg1}), we do not find statistically significant effects for misalignment ($p=0.075$) or eye occlusion interventions ($p=0.263$), but we find participants' accuracy on inverted authentic videos is 9.1 percentage points lower than when the videos are upright  ($p<0.001$).

The experimental results on non-recruited participants provide a replication and robustness check for the results on the recruited participants. The results from the non-recruited participants were not pre-registered because we weren't expecting many people to continue visiting our website organically. In fact, 9,188 visitors participated in the single video design between November 2020 and January 2021. In columns 4 through 6 in Table~\ref{fig:reg1}, we present the linear regressions results of the specialized face processing obstruction interventions on non-recruited participants' accuracy. Similar to the results for the recruited sample, we find statistically significant effects ($p<0.001$) of all three obstruction interventions on ability to accurately discern deepfakes from authentic videos. The number of observations in the non-recruited sample is over 16 times larger than the number of observations in the recruited sample. Likewise, the number of participants is 33 times larger. These numbers differ because the number of videos seen by participants in the non-recruited sample varied depending on participants' interest. With a larger sample size, we see statistically significant and negative effects of obstructions on all videos ranging from 4\% percentage point drop in accuracy from the eye occlusion intervention ($p<0.001$) to a 7\% drop on accuracy from the misalignment intervention ($p<0.001$). We also find all three treatments decrease participants' confidence scores by one half to one percentage point ($p<0.001$).

In the non-recruited sample, each of the 50 videos were viewed by between 945 to 1168 participants. We run separate linear regressions for each video and find statistically significant and negative effects at the 1\% significance level for inversion in 22 videos, misalignment in 14 videos, and occlusion in 22 videos. Furthermore, we find at least one of these specialized processing obstructions is negative and statistically significant at the 1\% significance for 29 of the 50 videos. 

In the sample of videos of political leaders, the specialized face processing obstructions had a significant effect on participants' ability to accurately identify the Vladimir Putin deepfake as a deepfake. The misalignment obstruction leads to a drop in accuracy of 20.7 percentage points ($p=0.001$). Likewise, the occlusion obstruction leads to a drop of 10.3 percentage points ($p=0.002$) and the inversion obstruction leads to a drop of 5.2 percentage points ($p=0.072$).

In columns 7 through 9 in Table~\ref{fig:reg1}, we present the linear regression results of the specialized face processing obstruction interventions on the model's predictions and find the computer vision model is affected by one specialized face processing obstruction but not the other two. We find the computer vision model's predictive accuracy drops by 12.1 percentage points on the inverted videos ($p=0.005$). We do not find a statistically significant difference in accuracy between either the control and misalignment sets of videos ($p=0.800$) or the control and occlusion sets of videos ($p=0.944$).

\begin{table*}[!htbp] \centering
\scalebox{0.73} {

\begin{tabular}{@{\extracolsep{5pt}}l|ccc|ccc|ccc}
\\[-1.8ex]\hline
\hline \\[-1.8ex]
& \multicolumn{9}{c}{\textit{Dependent variable: Accuracy}} \
\cr 
\\[-1.8ex] & \multicolumn{3}{c}{Recruited} & \multicolumn{3}{c}{Non-recruited} & \multicolumn{3}{c}{Computer}  \\
\\[-1.8ex] & All & Real & Fake & All & Real & Fake & All & Real & Fake \\
\hline \\[-1.8ex]
 Constant & 0.655$^{***}$ & 0.716$^{***}$ & 0.567$^{***}$ & 0.679$^{***}$ & 0.700$^{***}$ & 0.632$^{***}$ & 0.813$^{***}$ & 0.786$^{***}$ & 0.841$^{***}$ \\
  & (0.009) & (0.014) & (0.015) & (0.002) & (0.003) & (0.003) & (0.030) & (0.040) & (0.044) \\
 Inversion & -0.043$^{***}$ & -0.091$^{***}$ & 0.010$^{}$ & -0.053$^{***}$ & -0.080$^{***}$ & -0.027$^{***}$ & -0.121$^{***}$ & -0.110$^{*}$ & -0.132$^{**}$ \\
  & (0.014) & (0.021) & (0.021) & (0.004) & (0.006) & (0.006) & (0.042) & (0.056) & (0.063) \\
 Misalignment & -0.061$^{***}$ & -0.042$^{*}$ & -0.077$^{***}$ & -0.070$^{***}$ & -0.056$^{***}$ & -0.084$^{***}$ & 0.011$^{}$ & 0.000$^{}$ & 0.021$^{}$ \\
  & (0.016) & (0.024) & (0.025) & (0.005) & (0.007) & (0.007) & (0.042) & (0.056) & (0.063) \\
 Eye Occlusion & -0.044$^{***}$ & -0.023$^{}$ & -0.063$^{***}$ & -0.040$^{***}$ & -0.035$^{***}$ & -0.043$^{***}$ & -0.003$^{}$ & -0.007$^{}$ & 0.001$^{}$ \\
  & (0.015) & (0.021) & (0.024) & (0.004) & (0.006) & (0.006) & (0.042) & (0.056) & (0.063) \\
 \hline \\[-1.8ex]
 Anger & -0.020$^{}$ & -0.052$^{**}$ & 0.012$^{}$ & & & & & & \\
  & (0.014) & (0.024) & (0.021) & & & & & & \\
 \hline \\[-1.8ex]
 Number of Participants & 229 & 229 & 229 & 7563 & 6368 & 6670 & 0 & 0 & 0 \\
 Number of Guesses (Real) & 2349 & 1514 & 835 & 27446 & 18524 & 8922 & 81 & 76 & 5 \\
 Number of Guesses (Deepfake) & 1707 & 549 & 1158 & 22766 & 6316 & 16450 & 87 & 7 & 80 \\
 Number of Guesses (50-50) & 180 & 68 & 112 & 3713 & 1726 & 1987 & 32 & 17 & 15 \\
 Number of Unique Videos & 50 & 25 & 25 & 50 & 25 & 25 & 50 & 25 & 25 \\
\hline \\[-1.8ex]
 Observations & 4,236 & 2,131 & 2,105 & 53,925 & 26,566 & 27,359 & 200 & 100 & 100 \\
 $R^2$ & 0.180 & 0.069 & 0.225 & 0.185 & 0.057 & 0.273 & 0.062 & 0.054 & 0.073 \\
 Adjusted $R^2$ & 0.170 & 0.056 & 0.215 & 0.184 & 0.057 & 0.272 & 0.048 & 0.025 & 0.044 \\
 Residual Std. Error & 0.340 & 0.329 & 0.350 & 0.349 & 0.350 & 0.346 & 0.210 & 0.198 & 0.222  \\
 F Statistic & 288.686$^{***}$  & 164.804$^{***}$  & 169.388$^{***}$  & 3687.143$^{***}$  & 2150.874$^{***}$  & 4525.903$^{***}$  & 4.337$^{***}$  & 1.841$^{}$  & 2.514$^{*}$  \\
\hline
\hline \\[-1.8ex] 
\multicolumn{9}{l}{\textit{Note:} $^{*}$p$<$0.1; $^{**}$p$<$0.05; $^{***}$p$<$0.01} \\
\end{tabular}
}
\caption{Treatment effects of interventions on accuracy. Linear regressions on participant data includes video fixed effects with Eicker-Huber-White standard errors clustered at the participant level.}
\label{fig:reg1}

\end{table*}

\section*{Discussion}

How do ordinary human observers compare with the leading deepfake detection models? Our results are at odds with the commonly held view in media forensics that ordinary people have extremely limited ability to detect media manipulations. Past work in the cognitive science of media forensics has demonstrated that people are not good at perceiving and reasoning about shadow, reflection, and other physical implausibility cues~\cite{farid2010image, nightingale2019can, nightingale2017can, kasra2018seeing}. On first glance, deepfakes and other algorithmically generated images of people (e.g., images generated by StyleGAN) look quite realistic~\cite{karras2019style}. But, we show that deepfake algorithms generate artifacts that are perceptible to ordinary people, which may be partially explained by human's specialized visual processing of faces. In contrast to recent research showing ordinary people quickly learn to detect AI generated absences in photos~\cite{groh2019human}, we do not find evidence that participants improve in their ability to detect deepfakes.

By showing participants videos of unknown individuals making uncontroversial statements, we focused the truth discernment task specifically on visual perception. The lack of additional context creates a level playing field for a reasonably fair comparison of human and machine vision because humans cannot also reason about contextual, conceptual clues in these videos~\cite{firestone2020performance}. In the two alternative forced-choice paradigm of Experiment 1, 82\% of participants respond with higher accuracy than the leading model. In the more challenging single video framework in Experiment 2, participants still perform really well, and we find that between 13\% and 37\% of ordinary people outperform the leading deepfake detection model. When we aggregate participants' responses in Experiment 2, we find that collective intelligence, as measured by the crowd mean, is just as accurate as the model's prediction.

In the extension of the experiment to videos of well-known political leaders (Vladimir Putin and Kim Jong-un), participants significantly outperform the leading model, which is likely explained by participants' ability to go beyond visual perception of faces. Unlike the 50 sample holdout videos, participants could critically contemplate the authenticity of the video of the political leader. For example, participants might have considered whether Vladimir Putin or Kim Jong-un speak English, whether they actually sound like they do in the video, and whether such a well-known political figure would say such a thing. Not only do the majority of participants identify the deepfake status of videos of political leaders correctly, but the computer vision model is confident in its wrong predictions. Perhaps, the model failed because it was trained on face swapping deepfake manipulations as opposed to synthetic lip syncing manipulations. What the evidence shows is that today's leading model does not generalize well to stylistically different videos than the videos on which it has been trained, whereas human deepfake detection abilities do generalize across these different contexts. 

The model's predictions helped participants improve their accuracy overall, but whether a participant's accuracy increased depended on whether the model accurately identified the video as a deepfake or not. Participants often made significant adjustments based on the model's predictions, and inaccurate or equivocal model predictions led participants astray in 8 of 10 instances. Moreover, the model's incorrect assessment of the political leader deepfake videos is associated with a decrease in participant accuracy, which is in line with recent empirical research that shows deepfake warnings do not improve discernment of political videos~\cite{ternovski2021deepfake}. Likewise, these results mirror other recent research revealing human-AI collaborative decision making does not necessarily lead to more accurate results than either humans or AI alone~\cite{vaccaro2019effects,tschandl_humancomputer_2020, gaube2021ai, abeliuk2020quantifying, jacobs2021machine}.

Videos are heterogeneous, high-dimensional media, and as a result, participants were accurate on some videos on which the leading model failed and vice versa. In line with recent research examining perceptual differences between authentic and deepfake videos~\cite{wohler2021towards}, we identified 7 salient dimensions across the 50 sampled holdout videos to evaluate differences in how participants and the leading model discern authenticity: We find that the leading model performs slightly better than participants on low-quality videos that were categorized as grainy, blurry, and very dark. This differential performance suggests that the model is picking up on low-level details that participants appear to ignore. On the other hand, we find both recruited and non-recruited participants attain similar accuracies as the model on standard quality videos. Both participants and the model are quite adept at picking up on flickering faces. The model has trouble discerning between real and deepfake videos when two actors appear in the video while participants have no trouble in this context. This suggests that the model may be vulnerable to changes in context whereas participants are more robust to varying context. With respect to visual distractions, we find distractions are associated with participants identifying videos more often as deepfakes. While we showed recruited participants examples of distraction videos that should not be reported as deepfakes and we explicitly described these distractions in the instructions as not necessarily characteristic of deepfakes, we imagine the results concerning the distraction videos may possibly reflect confusion by the participants. Nonetheless, all reported results are robust to the exclusion of distraction videos. In light of recent research showing intersectional disparities in accuracy of commercial facial recognition software~\cite{buolamwini2018gender} and the impact of race on credibility with deepfakes~\cite{haut2021could}, we examine accuracy on the videos with dark skin actors. Participants and the leading model are both less accurate on videos with dark skin actors, but as we reported in the results section, we only find a statistically significant difference in participants' performance not the model's performance.

In Experiment 2, we find some evidence for our pre-registered hypothesis that anger would impair participants' ability to identify manipulated media. When we elicit incidental anger (i.e., anger unrelated to the task at hand), participants' accuracy at identifying real videos decreases, a pattern that held across almost all videos (see Figure S3 in the Supporting Information where participants assigned to the anger elicitation under perform participants assigned to the control in 22 out of 25 real videos, and see the Limitations section below). The negative and heterogeneous effect of incidental anger on the discernment of real (but not fake) videos may be related to the negative and heterogeneous effect of emotion priming on accuracy ratings of fake (but not real) news headlines~\cite{martel2019reliance}. Drawing on Martel et al 2020, one potential explanation for the negative effects of anger elicitation on the discernment of authentic but not deepfake videos is emotion leading to an over-reliance on intuition; in this experiment, if a participant sees something that looks like a deepfake manipulation, then she is unlikely to think the video is real, but if a participant does not see something that looks like a deepfake manipulation, then he might think he's simply unable to spot the detailed manipulation and may respond based on his intuition that a video is fake rather than whether he clearly saw a manipulation or not.

Both Experiments 1 and 2 provide support for the claim that specialized processing of faces helps people discern authenticity in visual media. In particular, we show that three visual obstructions designed to hinder specialized processing – inversion, misalignment, and partial occlusion – decrease participants' accuracy. In contrast to human visual processing, we find only inversion and not misalignment or partial occlusion change the model's performance. While the computer vision model is robust to misalignment and occlusion, this robustness may be a bug – the model overfitting to the training data – rather than a feature. Future research should explore whether specialized processing in computer vision models for deepfake detection enables better generalization to new contexts.

\section*{Limitations}

We evaluated human and machine performance on 167 videos (84 deepfake and 83 authentic videos) across Experiments 1 and 2. While these videos represent a balanced group of individuals across demographic dimensions and a variety of deepfake models, only the two political deepfake videos include lip syncing manipulations, which are some of the most commonly used models for producing political deepfakes~\cite{dolhansky2020deepfake, agarwal2019protecting,suwajanakorn2017synthesizing, prajwal_lip_2020}. Moreover, we do not specifically recruit expert fact-checkers or expert media forensic analysts, and as such, our results only generalize to the performance of ordinary people. Our comparison of untrained participants' predictions to the predictions of the leading computer vision model is limited to the best performance in 2020. If current trends continue as we expect they would, computer vision detection models will continue to improve (and possibly incorporate more human-like specialized processing of faces to better generalize across contexts) just as the realism of synthetic media generation algorithms will continue to improve. As a consequence, society will require more than just visual-based classification algorithms to protect against the potentially harmful threats that deepfakes pose~\cite{mirsky2020creation}.

The minimal context videos used here may not resemble the most problematic deepfakes because the videos here show unknown people saying non-controversial things in nondescript settings. On one hand, this minimal context makes the human participants' performance all the more impressive because such videos are missing many of the contextual cues they might normally use to discern authentic videos from deepfakes. On the other hand, perhaps videos designed to deceive are stylistically very different than the videos from the sampled holdout. As such, persuasive, manipulated video is important to consider in future research. The role of persuasion in synthetic media is beginning to be explored across varying media modalities~\cite{wittenberg_minimal_2020, dobber2020microtargeted}, but it is not the central focus of this paper. Instead, we ask how well the human visual processing system can detect the visual manipulations characteristic of deepfakes. We limit the bulk of our evaluation to uncontroversial videos of unknown actors to focus on the visual component of truth discernment. We begin to examine more realistic examples based on four videos of political leaders, but a larger sample size and further experimentation is necessary before making conclusions about how people judge the authenticity of political deepfakes. Furthermore, there is still much to learn about how AI systems and ordinary people can incorporate all the other information beyond facial features to make accurate judgments about a video's authenticity.

In this experiment, half of the videos were real and half deepfakes. This is useful for comparing human and machine performance, but this base rate of deepfakes does not reflect the base rate of misinformation in today's media ecosystems~\cite{allen2020evaluating}. In 2021, less than a fraction of a percent of news was misinformation~\cite{watts2021measuring}. Future experiments might consider examining people's ability to identify deepfakes when they do not have foreknowledge of the base rate of deepfakes. Moreover, an experiment embedded in a social media ecosystem could further identify how well people identify deepfakes within an ecologically valid context where people have access to contextual information such as who shared the video and how many others have shared or commented on the video. Ultimately, there are many ways to discern between real and fake videos, and visual perception should be considered as one tool in a user's toolkit for truth discernment.

We also considered how incidental emotions (i.e., emotions unrelated to the task at hand) affect participants' discernment of real and fake videos. Here, our two experiments found different results, and so we do not draw firm conclusions about the role of emotion on deepfake detection. In Experiment 1, the custom emotion elicitation interventions did not significantly alter deepfake detection performance --- though it also did not significantly alter self-reported emotions, making it unclear how much to read into the lack of effects on performance. The results from Experiment 2, though statistically significant by conventional standards, were near the cutoff for statistical significance for authentic videos and not statistically significant for deepfake videos. As such, future research could further explore the role of emotions in deepfake detection by running experiments with larger samples, examining additional emotions, ensuring effective elicitation, and focusing on integral emotions (emotions elicited directly from the stimuli). Recent research shows that inferences from feelings are context-sensitive and incidental emotions may be more likely to lead individuals astray in judgment tasks than integral emotions~\cite{schwarz2011feelings}.

\section*{Implications}

Relative to today's leading computer vision model, groups of individuals are just as accurate or more accurate depending on which videos are considered. Participants and the model perform equally well on standard resolution, visual-only deepfake manipulations. Participants perform better on the four political videos and attention check video while the computer vision model performs slightly better on blurry, grainy, and very dark videos. The model's poor performance on both deepfakes of world leaders and videos with two people instead of one suggests that the model may not generalize well to stylistically different videos than the videos on which it has been trained. Humans have no problem with this kind of generalization, and as a consequence, social media content moderation of video-based misinformation is likely to be more accurate when performed by teams of people than today's leading algorithm. As such, future research in crowd-based deepfake detection may consider how to most effectively aggregate wisdom of the crowds to improve discernment accuracy beyond the crowd mean (e.g., using algorithms such as the surprisingly popular answer~\cite{prelec_solution_2017} and revealed confidence~\cite{zhang2020identify}).

Sociotechnical systems may benefit from the combination of artificial intelligence and crowd-wisdom, but decision support tools for content moderation must be carefully designed to appropriately weigh human and model predictions. The confidently wrong predictions of the model on out-of-sample videos reveals the leading model is not ready to replace humans in detecting real-world deepfakes. Moreover, decision support tools can be counter-productive to accurate identification as evidenced by the many instances in which participants saw incorrect predictions from the model and subsequently adjusted their predictions to be less accurate. Instead of solely informing people on the likelihood that a model is a deepfake, crowd-wisdom could likely benefit from more explainable AI. Given that the leading model was more accurate at detecting certain classes of videos while humans were better at other classes, a future human-AI collaborative system might include additional information on video sub-types and how humans and machines perform across these sub-types. For example, video-level qualities (e.g., blurry, grainy, dark, specialized obstruction, stylistic similarities to training set, or other components upon which human and machine performance tends to diverge) and individual-level qualities could be factored into the interface and information presented by a human-AI collaborative system. By presenting model predictions alongside this information, it is possible humans could develop a better sense for confronting conflicting model predictions and deciding between second-guessing their own judgments and overriding the model's prediction. Machine-informed crowd-wisdom can be a promising approach to deepfake detection and other classification tasks more generally where human and machine classification performance is heterogeneous on sub-types of the data.

Specialized visual processing of faces helps humans discern between real and deepfake videos. In future instances when humans are tasked with deepfake detection, it is important to consider whether a video has been manipulated in such a way as to reduce specialized processing. Moreover, given the usefulness of specialized processing of faces for humans in detecting deepfakes, it is possible that computer vision models for deepfake detection may find use in incorporating (and/or learning) such specialized processing~\cite{farzmahdi2016specialized}. 

Visual cues will continue to be helpful in deepfake detection, but ultimately, identifying authentic video can involve much more than visual processing. When attempting to discern the truth from a lie, people rely on the available context, their knowledge of the world, their ability to critically reason, and their capacity to learn and update their beliefs. Similarly, the future of deepfake detection by both humans and machines should consider not only the perceptual clues but the greater context of a video and whether its message resembles an ordinary lie. 

\section*{Methods}

This research complies with all relevant ethical regulations and the Massachusetts Institute of Technology's Committee on the Use of Humans as Experimental Subjects determined this study to fall under Exempt Category 3 – Benign Behavioral Intervention. This study's exemption identification number is E-2070. All participants are informed that ``Detect Fakes is an MIT research project. All guesses will be collected for research purposes. All data for research is collected anonymously. For questions, please contact detectfakes@mit.edu. If you are under 18 years old, you need consent from your parents to use Deep Fakes.'' Most participants arrived at the website via organic links on the Internet. For recruited participants, we compensated each individual at a rate of \$7.28 an hour and provided bonus payments of 20\% to the top 10\% of participants. Before beginning the experiment, all recruited participants were also provided a research statement, ``The findings of this study are being used to shape science.  It is very important that you honestly follow the instructions requested of you on this task, which should take a total of 15 minutes. Check the box below based on your promise:'' with two options ``I promise to do the tasks with honesty and integrity, trying to do them uninterrupted with focus for the next 15 minutes.'' or ``I cannot promise this at this time.'' Participants who responded that they could not do this at this time were re-directed to the end of the experiment.

We hosted the experiment on a website called Detect Fakes at \url{https://detectfakes.media.mit.edu/}. Figure S4 in the Supporting Information presents a screenshot of the user interface for both Experiments 1 and 2. The rest of the methods are described in the Supplementary Information section. 

\section*{Data and Code Availability}

The datasets and code generated and analyzed during the current study are available in our public Github repository, \url{https://github.com/mattgroh/cognitive-science-detecting-deepfakes} (the Github repository will be set to public upon peer reviewed publication). All DFDC videos are available at \url{https://dfdc.ai/}~\cite{dolhansky2020deepfake} and the 5 non-DFDC videos are available in our public Github repository. 

\section*{Acknowledgements}

Authors would like to acknowledge funding from MIT Media Lab member companies, thank Alicia Guo for excellent research assistance, and thank the following communities for helpful feedback: the Affective Computing lab, the Perception and Mind lab, Human Cooperation lab, and the moderator and participants at the Human and AI Decision-Making panel at the CODE2020 conference. 

\bibliography{pnas-sample}

\section*{Supporting Information}

\subsection*{Experiment 1 – Two-Video Forced Choice}

\subsubsection*{Videos}

In the pilot experiment, we selected a sample of 56 pairs of videos from the competition training dataset of 19,154 authentic videos and 100,000 associated deepfakes. We selected a sample based on two criteria: First, we developed a neural network model to predict which videos are deepfakes and selected the 3\% of deepfake videos on which our model was highly confident it made the right decision but in fact was wrong. The goal of this selection process was to approximately identify the hardest videos for machines. Second, based on this subset of 3,000 deepfakes and 1,809 associated authentic videos, we randomly selected 56 unique deepfakes and their associated real video excluding obvious manipulations with fast flickering faces.

On the Detect Fakes website, two videos – one authentic and one deepfake manipulation of that authentic video – were displayed side-by-side. Videos were randomly assigned to be on the left or right side of the screen. Participants were asked to guess which video contains a deepfake manipulation. After each selection, participants were told which video was a deepfake and which was the original.

\subsubsection*{Participants}

5,524 individuals participated in this experiment. The sample size was pre-specified in the pre-analysis plan as the number of participants 3 months after the \nth{1000} participant engages in the experiment. Participants visited from all over the world and the top 5 most represented countries were United States (34\%), Germany (6\%), United Kingdom (5\%), Saudi Arabia (3\%), and Canada (3\%).

\subsubsection*{Experimental Design}

When participants visit the experiment, we store a cookie to keep track of each anonymous individual. We randomly assign participants to see one of 56 pairs of videos and allow participants to watch the videos and submit their guess as to which video is a deepfake. After participants submit their guess, we randomly assign another pair of videos without replacement. 

We cross-randomize participants to treatments within three interventions: (1) inversion, (2) emotion elicitation, and (3) reflection. In the inversion intervention, we randomly assign participants to see an inverted pair of videos on the \nth{4} or \nth{5} pair of videos that they see and show all participants a pair of inverted videos on the \nth{9} pair of videos shown. In the emotion elicitation intervention, we randomly assign participants to four control arms asking how happy, angry, anxious, or sad a participant feels and three treatment arms where we attempt to elicit incidental happiness, anger, and anxiety. In each control and treatment arm, we asked participants how happy, angry, anxious, or sad they feel on a scale of 1 to 10. The treatment arms include pop-up messages in between participants' guesses. The emotion elicitation intervention included a message, "There is scientific evidence that emotions help us make decisions. Think about a time you were really [angry/happy/anxious]. Before you continue, think about why you were really [angry/happy/anxious]." After each guess a similar message would pop up before participants would see the next videos. In the reflection intervention, we randomly assigned participants to a control group with normal load times and a treatment group where the website loads with a 600 millisecond delay.

\subsubsection*{Analysis}

We examine the causal effects of the three treatment assignments, $T_x$ on the accuracy score, $y_{i,j}$ of participant $i$ on manipulated video $j$. We include video fixed effects, $\mu_j$, and the following fixed-effects linear regression model reflects the pre-analysis plan with $\epsilon_{i,j}$ representing the error term:

\begin{equation}
y_{i,j} = \alpha + 
\beta_1 T_{1,i}  +
\beta_2 T_{2,i,j}  +
\beta_3 T_{3,i}  +
\mu_j + \epsilon_{i,j}
\label{equation:one}
\end{equation}

We cluster errors at the individual level with Eicker-Huber-White robust standard errors because the emotion elicitation assignment and reflection assignment is conducted at the participant level~\cite{abadie2017should}. As hypothesized in the pre-analysis plan, the inversion treatment has a negative effect ($p=0.004$). The emotion elicitation treatments did not appear to effectively elicit emotions, and as a result we do not find an effect ($p=0.209$ for anger, $p=0.867$ for happy, $p=0.560$ for anxiety). In fact, we do not find evidence that the self-reported emotional intensities of the participants assigned to the emotion elicitation treatment were different than the intensities reported by the control group ($p=0.190$ for happy, $p=0.558$ for anxiety, $p=0.091$ for anger). We interpret these results to indicate that our emotion elicitation intervention in Experiment 1 did not reliably elicit emotional responses. We do not find an effect of the reflection treatment on outcomes ($p=0.592$). 

\subsection*{Experiment 2 – Single Video Design}

\subsubsection*{Videos}

In Experiment 2, we randomly sampled 50 videos from the 4,000 videos in the DeepFakes Detection Competition holdout set~\cite{dolhansky2020deepfake}. 25 videos are deepfakes and 25 videos are authentic. Unlike videos from the competition training set sampled in the pilot experiment, some the these videos included distractions such as flickering shapes and faces that move around the video intended to throw off the computer vision model. In order to ensure participants did not confuse these distractions for deepfake manipulations, we explicitly described these distractions as not necessarily deepfakes in the instructions. In addition, we show several examples to recruited participants before the experiment begins. We do not show these examples to non-recruited participants because we wanted to increase engagement for people visiting the website on their own volition. We also included one attention check deepfake video and four videos of Vladimir Putin and Kim Jong-un (two real videos and two deepfakes each). All videos were 10 seconds long.

After completing the attention check, participants see a sequence of single videos that are randomly chosen from the remaining 54 videos and are randomly assigned to include a holistic obstruction manipulation. 

We generate predictions for the computer vision model by running the competition winning computer vision model on the 54 sampled videos and on the 162 inversion, misalignment, and occlusion transformations of these videos~\cite{selim2020}.

\subsubsection*{Participants}

We recruited 304 individuals from Prolific to participate in the pre-registered experiment. Participants were recruited from Prolific based on a nationally representative sample from the United States across age, ethnicity, and sex~\cite{palan_prolificacsubject_2018}. In addition, 9,188 non-recruited individuals participated. The sample size for the non-recruited sample was not pre-specified in the pre-analysis plan because we did not expect so many people to continue visiting the website. As such, we analyze the non-recruited participants in Experiment 2 in the spirit of the other two pre-analysis plans. Non-recruited participants visited from all over the world and the top 5 most represented countries were United States (31\%), Brazil (9\%), United Kingdom (7\%), Canada (4\%), and Germany (4\%).

\subsubsection*{Experimental Design}

\begin{figure*}[hb]
    \centering
    \includegraphics[width=0.9\textwidth]{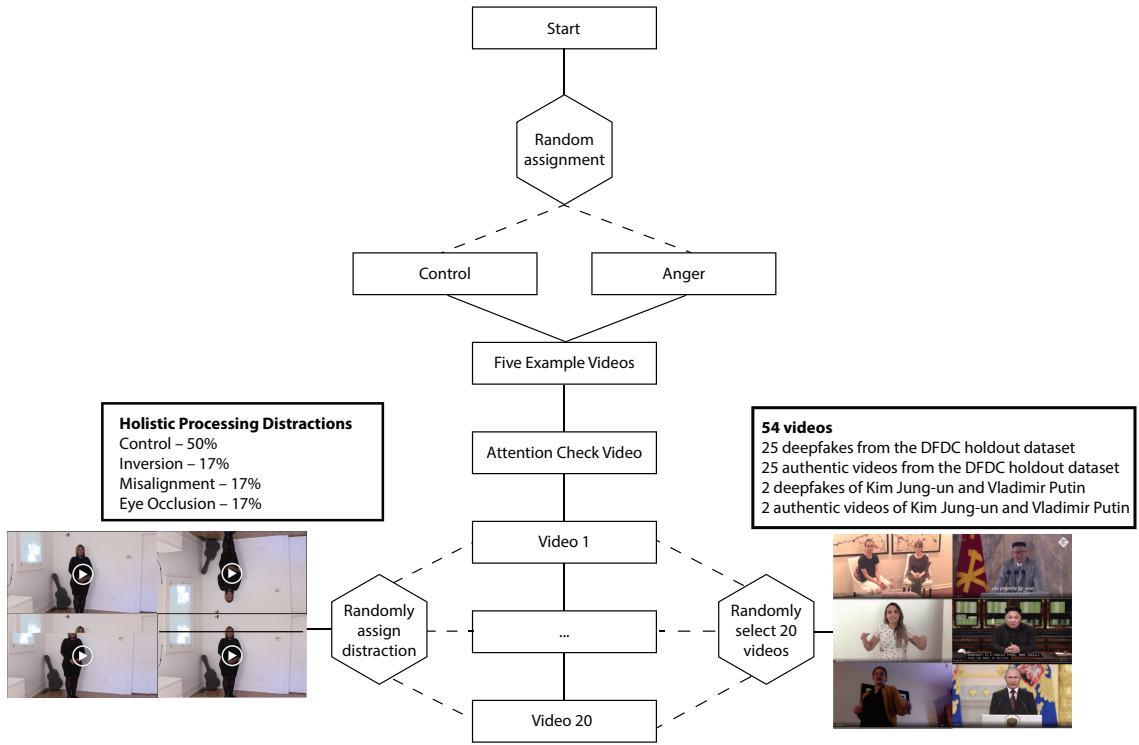}
    \caption{Recruited participants begin the experiment with a short writing assignment where participants either describe things that make them angry or things they did today. Then, we present five example videos to clarify to participants the difference between deepfakes, holistic processing distractions, and video distractions. Next, we present an attention check video to participants. Finally, we present videos from our stimuli set which are ordered randomly and are assigned to holistic processing distractions at random.}
    \label{fig:experimental_design}
\end{figure*}

We present the design flow for Experiment 2 in Figure~\ref{fig:experimental_design}. Recruited participants begin at the start while non-recruited participants begin at the attention check video. All participants see an instruction modal when they first visit the website.

First, participants are randomly assigned to either an incidental anger elicitation intervention or a control exercise. These exercises are based on asking participants to respond to two questions displayed in Appendix Table~\ref{fig:emotionelicitation}, which have been shown to elicit incidental emotions in research experiments~\cite{small2008emotional, castagnetti2020anger}.

Next, participants are shown five example videos to learn the difference between deepfakes and other distractions including both the flickering faces and the treatment interventions (inversion, misalignment, and occlusion). 

Next, all participants in Experiment 2 see an attention check video where a man's face clearly morphs from one configuration to another while he says, "This is a PSA. You can't believe anything you see these days. These glasses aren't even real. Neither is my face..." 

After participants view the attention check video, they are shown a single video randomly drawn from the sample of 54 videos. Videos are never shown twice and participants see just as many deepfakes as real videos. 50\% of videos are shown without a holistic processing obstruction. The other half are are shown upside down (1/6 of total observations), with the top and bottom half of the actor's face misaligned (1/6 of total observations), and one sixth are shown with the eyes occluded by a thin black strip (1/6 of total observations). Recruited participants are asked to view 20 videos while non-recruited participants can view up to 45 videos.

\subsubsection*{Analysis}

We examine the causal effects of the four treatment assignments, $T_x$ on the accuracy score (normalized for correctness), $y_{i,j}$ of participant $i$ on manipulated video $j$. We include video fixed effects, $\mu_j$, and the following fixed-effects linear regression model reflects the pre-analysis plan with $\epsilon_{i,j}$ representing the error term:

\begin{equation}
y_{i,j} = \alpha + 
\beta_1 T_{1,i}  +
\beta_2 T_{2,i,j}  +
\beta_3 T_{3,i,j}  +
\beta_4 T_{4,i,j}  +
\mu_j + \epsilon_{i,j}
\label{equation:two}
\end{equation}

We include video fixed effects to control for the differential accuracy across videos. We cluster errors at the individual level with Eicker-Huber-White robust standard errors because the emotion elicitation assignment is conducted at the participant level~\cite{abadie2017should}. As specified in the pre-analysis plan, all analyses exclude participants who fail this attention check. Nevertheless, the results in these experiments remain qualitatively the same when including these participants.

\begin{table*}[!htbp] \centering
\scalebox{0.95} {
\begin{tabular}{p{0.075\linewidth} | p{0.4\linewidth} | p{0.5\linewidth}}
\toprule
                Treatment & Question 1 & Question 2 \\
\midrule
             Anger &
             What are the three to five things that make you most angry? Please write two-three sentences about each thing that makes you angry. (Examples of things you might write about include: being treated unfairly by someone, being insulted or offended, etc.) &  
             Now, we’d like you to describe in more detail the one situation that makes you (or has made you) most angry. This could be something you are presently experiencing or something from the past. Begin by writing down what you remember of the anger-inducing event(s) and continue by writing as detailed a description of the event(s) as is possible. If you can, please write your description so that someone reading this might even get angry just from learning about the situation. What is it like to be in this situation? Why does it make you so angry?  \\
             \hline \\[-1.8ex]
             Control & What are three to five activities that you did today? Please write two-three sentences about each activity that you decide to share. (Examples of things you might write about include: walking, eating lunch, brushing your teeth, etc.) & Now, we’d like you to describe in more detail the way you typically spend your evenings. Begin by writing down a description of your activities and then figure out how much time you devote to each activity. Examples of things you might describe include eating dinner, studying for an exam, working, talking to friends, watching TV, etc. If you can, please write your description so that someone reading this might be able to reconstruct the way in which you, specifically, spend your evenings.  \\
             \bottomrule
\end{tabular}
}
\caption{Emotion elicitation questions adapted from Lerner and Small 2008~\cite{small2008emotional}}.
\label{fig:emotionelicitation}
\end{table*}

\begin{table*}[!htbp] \centering
\scalebox{0.85} {
\begin{tabular}{lrrrr}
\toprule
                Video &  Crowd Mean (Recruited) &  Crowd Mean (Non-recruited) &  Computer Vision &  Fake \\
\midrule
            7837 &                      0.31 &                      0.36 &             0.23 &     1 \\
             5843 &                      0.33 &                      0.29 &             0.99 &     1 \\
             4555 &                      0.35 &                      0.21 &             0.50 &     1 \\
             5319 &                      0.35 &                      0.51 &             0.99 &     1 \\
             4757 &                      0.37 &                      0.53 &             0.54 &     0 \\
             5471 &                      0.39 &                      0.46 &             0.99 &     1 \\
             4110 &                      0.44 &                      0.46 &             0.75 &     1 \\
             5142 &                      0.47 &                      0.48 &             0.62 &     1 \\
             7298 &                      0.48 &                      0.48 &             0.63 &     1 \\
             5785 &                      0.49 &                      0.49 &             0.99 &     1 \\
             4545 &                      0.49 &                      0.73 &             0.93 &     0 \\
             6660 &                      0.49 &                      0.44 &             0.96 &     1 \\
             7693 &                      0.53 &                      0.65 &             0.99 &     1 \\
        Jong-un (Deepfake) &                      0.53 &                      0.71 &             0.02 &     1 \\
             6537 &                      0.53 &                      0.41 &             0.98 &     1 \\
             6642 &                      0.54 &                      0.78 &             0.99 &     1 \\
        Jong-un (Real) &                      0.55 &                      0.72 &             0.55 &     0 \\
             7679 &                      0.59 &                      0.73 &             0.99 &     1 \\
      Putin (Real) &                      0.63 &                      0.66 &             0.99 &     0 \\
             7968 &                      0.63 &                      0.80 &             0.50 &     1 \\
             6597 &                      0.64 &                      0.56 &             0.99 &     0 \\
             6251 &                      0.65 &                      0.69 &             0.98 &     0 \\
             6564 &                      0.66 &                      0.79 &             0.76 &     1 \\
             6561 &                      0.66 &                      0.66 &             0.50 &     0 \\
             4704 &                      0.66 &                      0.78 &             0.98 &     1 \\
             4261 &                      0.67 &                      0.70 &             0.80 &     0 \\
             6153 &                      0.67 &                      0.89 &             0.30 &     1 \\
             7356 &                      0.68 &                      0.69 &             0.50 &     0 \\
             5370 &                      0.70 &                      0.78 &             0.97 &     0 \\
             5844 &                      0.71 &                      0.66 &             0.28 &     0 \\
             4122 &                      0.71 &                      0.72 &             0.99 &     0 \\
             5534 &                      0.71 &                      0.54 &             0.56 &     0 \\
             7723 &                      0.71 &                      0.69 &             0.89 &     0 \\
             6651 &                      0.71 &                      0.78 &             0.99 &     0 \\
      Putin (Deepfake) &                      0.74 &                      0.83 &             0.08 &     1 \\
             7737 &                      0.74 &                      0.78 &             0.97 &     0 \\
             6676 &                      0.75 &                      0.83 &             0.97 &     0 \\
             4250 &                      0.76 &                      0.87 &             0.98 &     1 \\
             4712 &                      0.76 &                      0.75 &             0.99 &     0 \\
             7324 &                      0.77 &                      0.79 &             0.69 &     0 \\
             7266 &                      0.78 &                      0.93 &             0.99 &     1 \\
             5908 &                      0.79 &                      0.70 &             0.97 &     0 \\
             6216 &                      0.79 &                      0.84 &             0.99 &     1 \\
             7494 &                      0.80 &                      0.82 &             0.91 &     0 \\
             7439 &                      0.80 &                      0.93 &             0.99 &     1 \\
             6575 &                      0.81 &                      0.95 &             0.98 &     1 \\
             6189 &                      0.82 &                      0.67 &             0.89 &     0 \\
 Attention Check &                      0.82 &                      0.86 &             0.01 &     1 \\
             7849 &                      0.82 &                      0.81 &             0.50 &     0 \\
             7957 &                      0.84 &                      0.79 &             0.84 &     0 \\
             4135 &                      0.85 &                      0.83 &             0.50 &     0 \\
             5201 &                      0.88 &                      0.83 &             0.99 &     0 \\
             4121 &                      0.88 &                      0.98 &             0.99 &     1 \\
             6592 &                      0.89 &                      0.84 &             0.50 &     0 \\
             4742 &                      0.89 &                      0.95 &             0.99 &     1 \\
\bottomrule
\end{tabular}
}
\caption{Videos from Experiment 2 and Accuracy Scores}
\label{fig:videoaccuracies}
\end{table*}

\begin{table*}[!htbp] \centering

\begin{tabular}{cc|c|c|c|}
\textbf{E2 Control Videos – Recruited} &\multicolumn{1}{c}{}&\multicolumn{2}{c}{Predicted Class}\\
&\multicolumn{1}{c}{}&\multicolumn{1}{c}{Fake}
&\multicolumn{1}{c}{Real}
&\multicolumn{1}{c}{\textbf{}}\\
\cline{3-5}
\multicolumn{1}{c}{\multirow{4}{*}{{Actual Class}}}
& {\multirow{2}{*}{{Fake}}} & {\multirow{2}{*}{{792}}} & {\multirow{2}{*}{{525}}} & Sensitivity \\
&  &  & & 60\% \\
\cline{3-5}
& {\multirow{2}{*}{{Real}}} & {\multirow{2}{*}{{333}}} & {\multirow{2}{*}{{984}}} & Specificity \\
&  &  & & 75\% \\
\cline{3-5}
& {\multirow{2}{*}{{}}} & Precision & NPV & Accuracy \\
&  & 70\% & 65\% & 67\%  \\\cline{3-5}
\vspace{.2cm}
\end{tabular}

\begin{tabular}{cc|c|c|c|}
\textbf{E2 Control Videos – Non-recruited} &\multicolumn{1}{c}{}&\multicolumn{2}{c}{Predicted Class}\\
&\multicolumn{1}{c}{}&\multicolumn{1}{c}{Fake}
&\multicolumn{1}{c}{Real}
&\multicolumn{1}{c}{\textbf{}}\\
\cline{3-5}
\multicolumn{1}{c}{\multirow{4}{*}{{Actual Class}}}
& {\multirow{2}{*}{{Fake}}} & {\multirow{2}{*}{{8,690}}} & {\multirow{2}{*}{{4,140}}} & Sensitivity \\
&  &  & & 68\% \\
\cline{3-5}
& {\multirow{2}{*}{{Real}}} & {\multirow{2}{*}{{2,824}}} & {\multirow{2}{*}{{9,814}}} & Specificity \\
&  &  & & 78\% \\
\cline{3-5}
& {\multirow{2}{*}{{}}} & Precision & NPV & Accuracy \\
&  & 75\% & 70\% & 73\% \\\cline{3-5}
\end{tabular}

\caption{Confusion matrices for participant accuracy on control videos in Experiment 2. These confusion matrices exclude the 186 and 1,795 guesses made by participants (in the recruited and non-recruited cohorts, respectively) who guessed 50-50 because such guesses can neither be classified as predicting real or fake. NPV stands for negative predictive value.}
\label{fig:confusion_0}
\end{table*}

\begin{table*}[!htbp] \centering

\begin{tabular}{cc|c|c|c|}
\textbf{E2 Eye Occlusion Videos – Recruited} &\multicolumn{1}{c}{}&\multicolumn{2}{c}{Predicted Class}\\
&\multicolumn{1}{c}{}&\multicolumn{1}{c}{Fake}
&\multicolumn{1}{c}{Real}
&\multicolumn{1}{c}{\textbf{}}\\
\cline{3-5}
\multicolumn{1}{c}{\multirow{4}{*}{{Actual Class}}}
& {\multirow{2}{*}{{Fake}}} & {\multirow{2}{*}{{250}}} & {\multirow{2}{*}{{210}}} & Sensitivity \\
&  &  & & 54\% \\
\cline{3-5}
& {\multirow{2}{*}{{Real}}} & {\multirow{2}{*}{{129}}} & {\multirow{2}{*}{{337}}} & Specificity \\
&  &  & & 72\% \\
\cline{3-5}
& {\multirow{2}{*}{{}}} & Precision & NPV & Accuracy \\
&  & 66\% & 62\% & 63\%  \\\cline{3-5}
\vspace{.2cm}
\end{tabular}

\begin{tabular}{cc|c|c|c|}
\textbf{E2 Eye Occlusion Videos – Non-recruited} &\multicolumn{1}{c}{}&\multicolumn{2}{c}{Predicted Class}\\
&\multicolumn{1}{c}{}&\multicolumn{1}{c}{Fake}
&\multicolumn{1}{c}{Real}
&\multicolumn{1}{c}{\textbf{}}\\
\cline{3-5}
\multicolumn{1}{c}{\multirow{4}{*}{{Actual Class}}}
& {\multirow{2}{*}{{Fake}}} & {\multirow{2}{*}{{2,652}}} & {\multirow{2}{*}{{1,564}}} & Sensitivity \\
&  &  & & 63\% \\
\cline{3-5}
& {\multirow{2}{*}{{Real}}} & {\multirow{2}{*}{{1,076}}} & {\multirow{2}{*}{{3,037}}} & Specificity \\
&  &  & & 74\% \\
\cline{3-5}
& {\multirow{2}{*}{{}}} & Precision & NPV & Accuracy \\
&  & 71\% & 66\% & 68\% \\\cline{3-5}
\end{tabular}

\caption{Confusion matrices for participant accuracy on eye occlusion videos in Experiment 2. These confusion matrices exclude the 62 and 638 guesses made by participants (in the recruited and non-recruited cohorts, respectively) who guessed 50-50 because such guesses can neither be classified as predicting real or fake. NPV stands for negative predictive value.}
\label{fig:confusion_1}
\end{table*}

\begin{table*}[!htbp] \centering

\begin{tabular}{cc|c|c|c|}
\textbf{E2 Misalignment Videos – Recruited} &\multicolumn{1}{c}{}&\multicolumn{2}{c}{Predicted Class}\\
&\multicolumn{1}{c}{}&\multicolumn{1}{c}{Fake}
&\multicolumn{1}{c}{Real}
&\multicolumn{1}{c}{\textbf{}}\\
\cline{3-5}
\multicolumn{1}{c}{\multirow{4}{*}{{Actual Class}}}
& {\multirow{2}{*}{{Fake}}} & {\multirow{2}{*}{{211}}} & {\multirow{2}{*}{{213}}} & Sensitivity \\
&  &  & & 50\% \\
\cline{3-5}
& {\multirow{2}{*}{{Real}}} & {\multirow{2}{*}{{123}}} & {\multirow{2}{*}{{286}}} & Specificity \\
&  &  & & 70\% \\
\cline{3-5}
& {\multirow{2}{*}{{}}} & Precision & NPV & Accuracy \\
&  & 63\% & 57\% & 60\%  \\\cline{3-5}
\vspace{.2cm}
\end{tabular}

\begin{tabular}{cc|c|c|c|}
\textbf{E2 Misalignment Videos – Non-recruited} &\multicolumn{1}{c}{}&\multicolumn{2}{c}{Predicted Class}\\
&\multicolumn{1}{c}{}&\multicolumn{1}{c}{Fake}
&\multicolumn{1}{c}{Real}
&\multicolumn{1}{c}{\textbf{}}\\
\cline{3-5}
\multicolumn{1}{c}{\multirow{4}{*}{{Actual Class}}}
& {\multirow{2}{*}{{Fake}}} & {\multirow{2}{*}{{2,427}}} & {\multirow{2}{*}{{1,779}}} & Sensitivity \\
&  &  & & 58\% \\
\cline{3-5}
& {\multirow{2}{*}{{Real}}} & {\multirow{2}{*}{{1,135}}} & {\multirow{2}{*}{{2,932}}} & Specificity \\
&  &  & & 72\% \\
\cline{3-5}
& {\multirow{2}{*}{{}}} & Precision & NPV & Accuracy \\
&  & 68\% & 62\% & 65\% \\\cline{3-5}
\end{tabular}

\caption{Confusion matrices for participant accuracy on misalignment videos in Experiment 2. These confusion matrices exclude the 60 and 627 guesses made by participants (in the recruited and non-recruited cohorts, respectively) who guessed 50-50 because such guesses can neither be classified as predicting real or fake. NPV stands for negative predictive value.}
\label{fig:confusion_2}
\end{table*}

\begin{table*}[!htbp] \centering

\begin{tabular}{cc|c|c|c|}
\textbf{E2 Inverted Videos – Recruited} &\multicolumn{1}{c}{}&\multicolumn{2}{c}{Predicted Class}\\
&\multicolumn{1}{c}{}&\multicolumn{1}{c}{Fake}
&\multicolumn{1}{c}{Real}
&\multicolumn{1}{c}{\textbf{}}\\
\cline{3-5}
\multicolumn{1}{c}{\multirow{4}{*}{{Actual Class}}}
& {\multirow{2}{*}{{Fake}}} & {\multirow{2}{*}{{245}}} & {\multirow{2}{*}{{164}}} & Sensitivity \\
&  &  & & 60\% \\
\cline{3-5}
& {\multirow{2}{*}{{Real}}} & {\multirow{2}{*}{{163}}} & {\multirow{2}{*}{{296}}} & Specificity \\
&  &  & & 64\% \\
\cline{3-5}
& {\multirow{2}{*}{{}}} & Precision & NPV & Accuracy \\
&  & 60\% & 64\% & 62\%  \\\cline{3-5}
\vspace{.2cm}
\end{tabular}

\begin{tabular}{cc|c|c|c|}
\textbf{E2 Inverted Videos – Non-recruited} &\multicolumn{1}{c}{}&\multicolumn{2}{c}{Predicted Class}\\
&\multicolumn{1}{c}{}&\multicolumn{1}{c}{Fake}
&\multicolumn{1}{c}{Real}
&\multicolumn{1}{c}{\textbf{}}\\
\cline{3-5}
\multicolumn{1}{c}{\multirow{4}{*}{{Actual Class}}}
& {\multirow{2}{*}{{Fake}}} & {\multirow{2}{*}{{2,688}}} & {\multirow{2}{*}{{1,440}}} & Sensitivity \\
&  &  & & 65\% \\
\cline{3-5}
& {\multirow{2}{*}{{Real}}} & {\multirow{2}{*}{{1,282}}} & {\multirow{2}{*}{{2,2745}}} & Specificity \\
&  &  & & 68\% \\
\cline{3-5}
& {\multirow{2}{*}{{}}} & Precision & NPV & Accuracy \\
&  & 68\% & 66\% & 67\% \\\cline{3-5}
\end{tabular}

\caption{Confusion matrices for participant accuracy on inverted videos in Experiment 2. These confusion matrices exclude the 65 and 653 guesses made by participants (in the recruited and non-recruited cohorts, respectively) who guessed 50-50 because such guesses can neither be classified as predicting real or fake. NPV stands for negative predictive value.}
\label{fig:confusion_3}
\end{table*}

\begin{table*}[!htbp] \centering

\begin{tabular}{cc|c|c|c|}
\textbf{E2 Emotion Condition – Control} &\multicolumn{1}{c}{}&\multicolumn{2}{c}{Predicted Class}\\
&\multicolumn{1}{c}{}&\multicolumn{1}{c}{Fake}
&\multicolumn{1}{c}{Real}
&\multicolumn{1}{c}{\textbf{}}\\
\cline{3-5}
\multicolumn{1}{c}{\multirow{4}{*}{{Actual Class}}}
& {\multirow{2}{*}{{Fake}}} & {\multirow{2}{*}{{799}}} & {\multirow{2}{*}{{612}}} & Sensitivity \\
&  &  & & 57\% \\
\cline{3-5}
& {\multirow{2}{*}{{Real}}} & {\multirow{2}{*}{{363}}} & {\multirow{2}{*}{{1,082}}} & Specificity \\
&  &  & & 75\% \\
\cline{3-5}
& {\multirow{2}{*}{{}}} & Precision & NPV & Accuracy \\
&  & 69\% & 64\% & 66\%  \\\cline{3-5}
\vspace{.2cm}
\end{tabular}

\begin{tabular}{cc|c|c|c|}
\textbf{E2 Emotion Condition – Anger} &\multicolumn{1}{c}{}&\multicolumn{2}{c}{Predicted Class}\\
&\multicolumn{1}{c}{}&\multicolumn{1}{c}{Fake}
&\multicolumn{1}{c}{Real}
&\multicolumn{1}{c}{\textbf{}}\\
\cline{3-5}
\multicolumn{1}{c}{\multirow{4}{*}{{Actual Class}}}
& {\multirow{2}{*}{{Fake}}} & {\multirow{2}{*}{{699}}} & {\multirow{2}{*}{{500}}} & Sensitivity \\
&  &  & & 58\% \\
\cline{3-5}
& {\multirow{2}{*}{{Real}}} & {\multirow{2}{*}{{385}}} & {\multirow{2}{*}{{821}}} & Specificity \\
&  &  & & 68\% \\
\cline{3-5}
& {\multirow{2}{*}{{}}} & Precision & NPV & Accuracy \\
&  & 64\% & 62\% & 63\% \\\cline{3-5}
\end{tabular}

\caption{Confusion matrices for participant accuracy based on assignment to emotion elicitation conditions in Experiment 2. These confusion matrices exclude the 218 and 155 guesses made by participants (in the control and anger elicitation cohorts, respectively) who guessed 50-50 because such guesses can neither be classified as predicting real or fake. NPV stands for negative predictive value.}
\label{fig:confusion_4}
\end{table*}

\begin{figure*}[ht]
    \centering
    \includegraphics[width=0.49\textwidth]{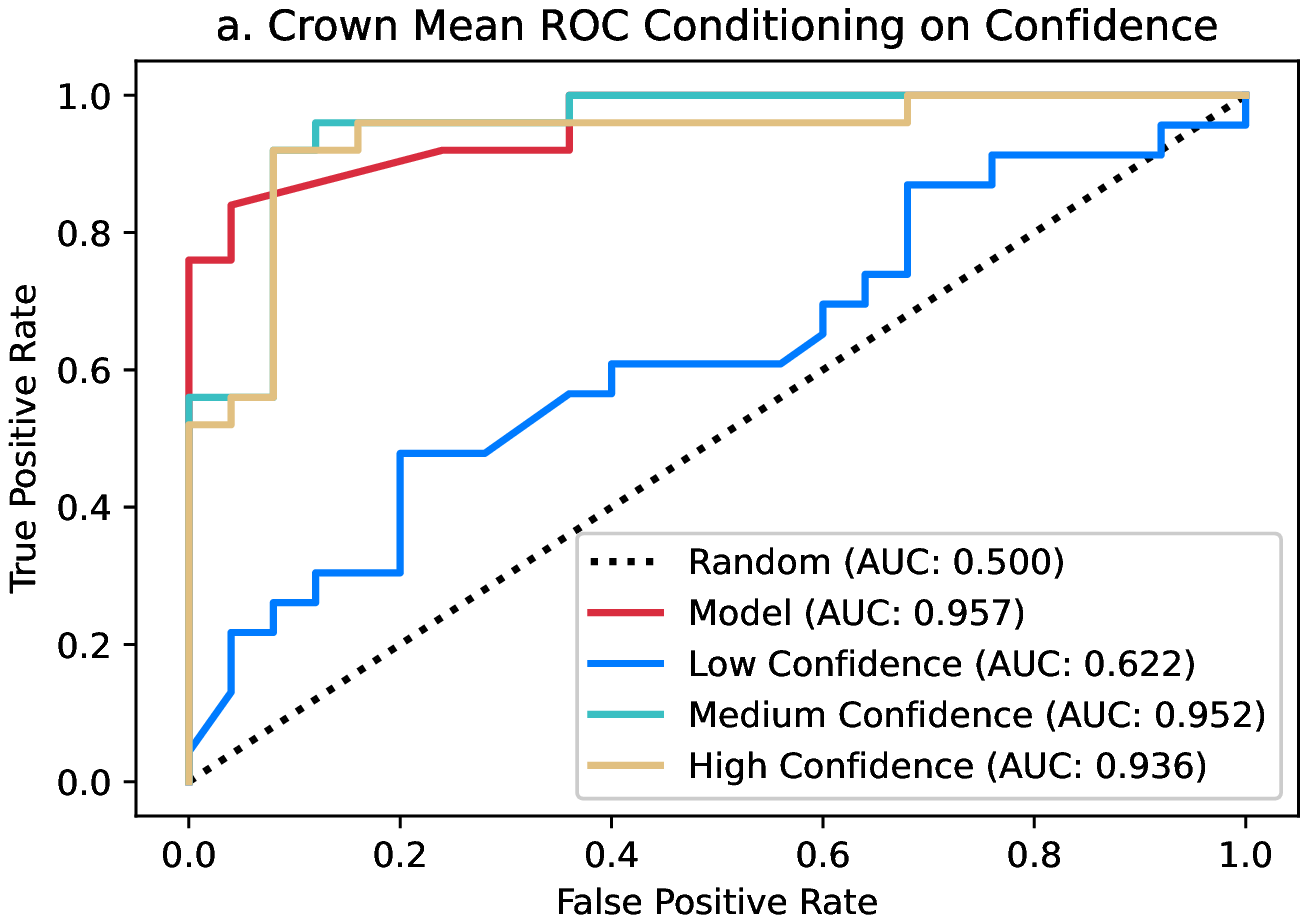}
    \includegraphics[width=0.49\textwidth]{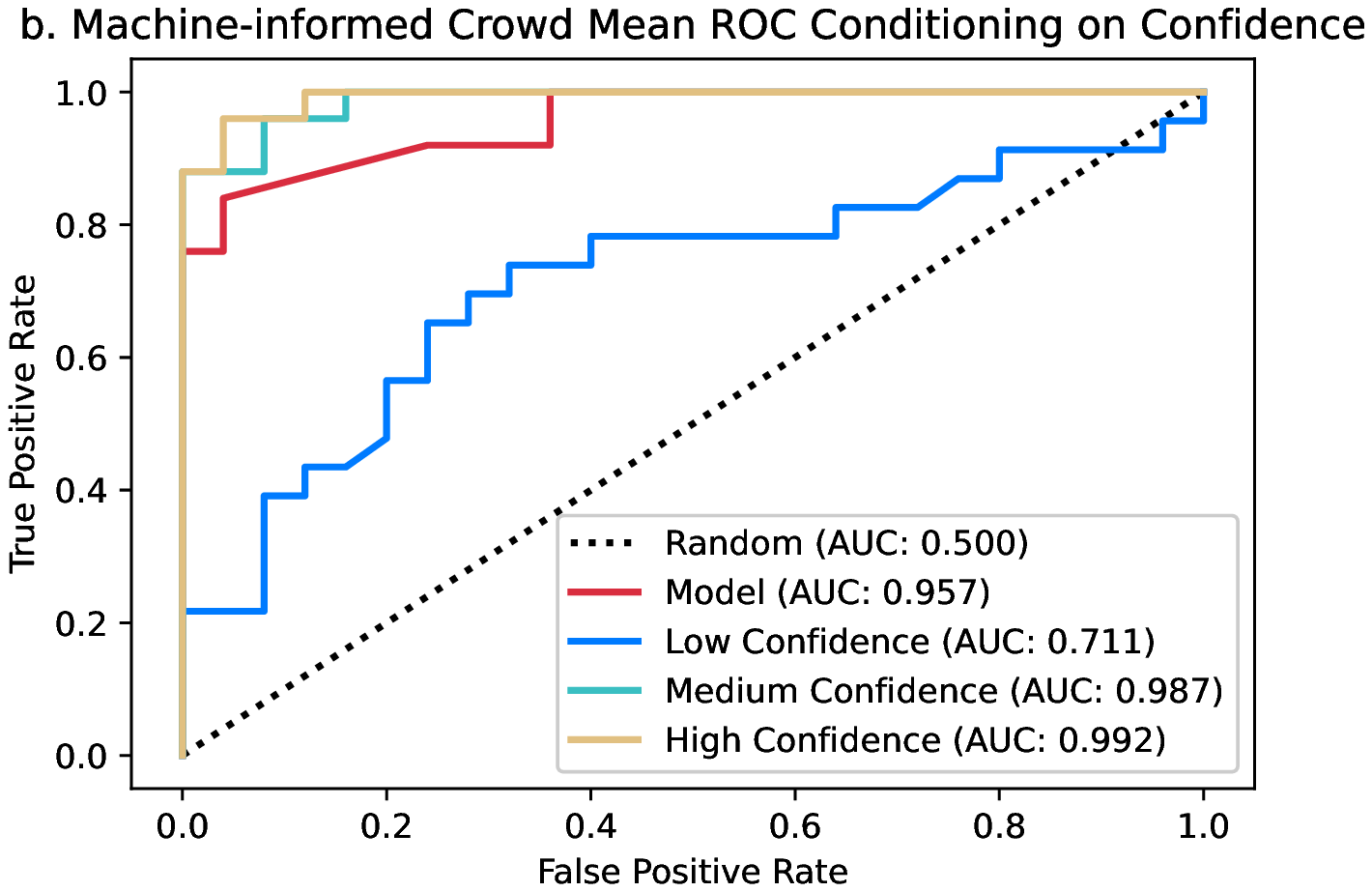}
    \caption{Figures~\ref{fig:coc}a and ~\ref{fig:coc}b present the receiver operator characteristic curves of participants' collective performance conditioned on their confidence. Low confidence is defined as responses between 33.5 and 66.5, medium confidence is defined as responses between 17 and 33.5 or 66.5 and 83, and high confidence is defined as responses between 0 and 17 or 83 and 100.}
    \label{fig:coc}
\end{figure*}

\begin{figure*}[ht]
    \centering
    \includegraphics[width=0.49\textwidth]{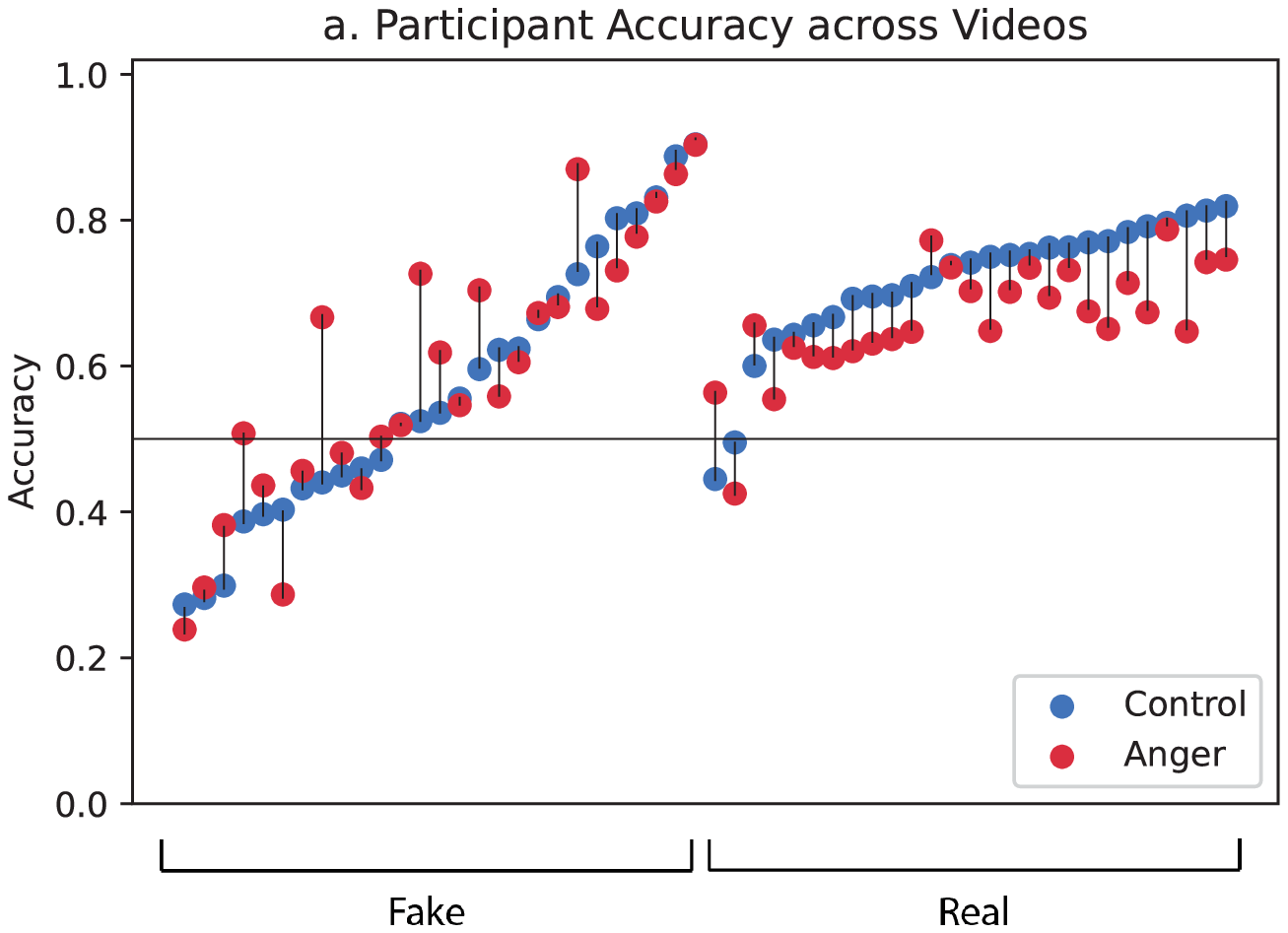}
    \includegraphics[width=0.49\textwidth]{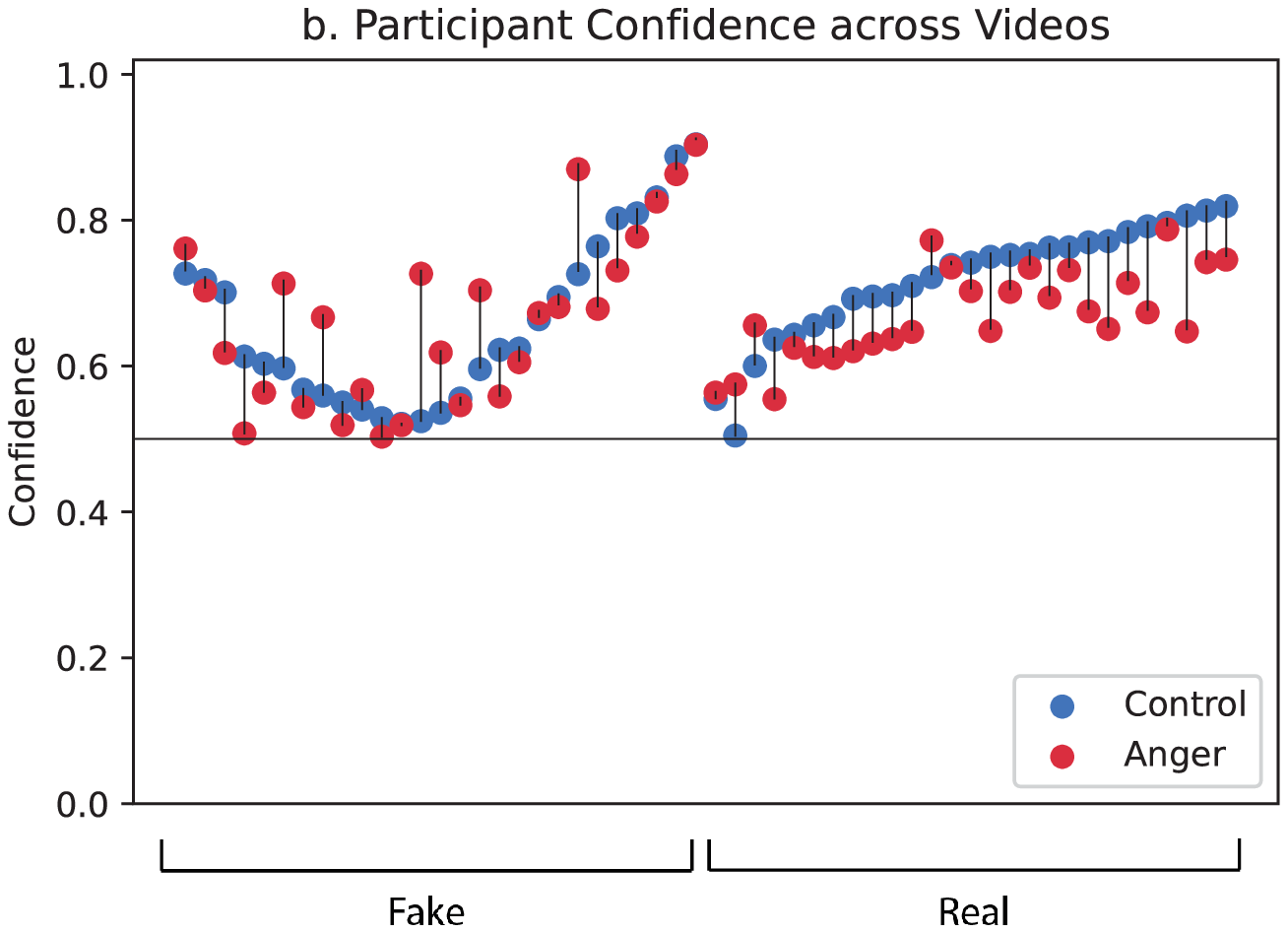}
    \caption{The mean accuracy and confidence scores on each video are plotted against assignment to the emotion elicitation conditions. In Figure~\ref{fig:anger}a, accuracy is defined as the participant's response between 0 and 1 normalized for correctness, which is the participant's response if correct or 1 minus the participant's response if incorrect. In Figure~\ref{fig:anger}b, confidence is defined orthogonal to correctness as 0.5 plus the absolute value of the participants' response minus 0.5 (e.g., a response of .49 is transformed to .51; a response of .02 is transformed to .98). The anger elicitation condition leads participants to reduce both accuracy and reported confidence when examining real videos.}
    \label{fig:anger}
\end{figure*}

\begin{figure*}[ht]
    \centering
    \includegraphics[width=0.85\textwidth]{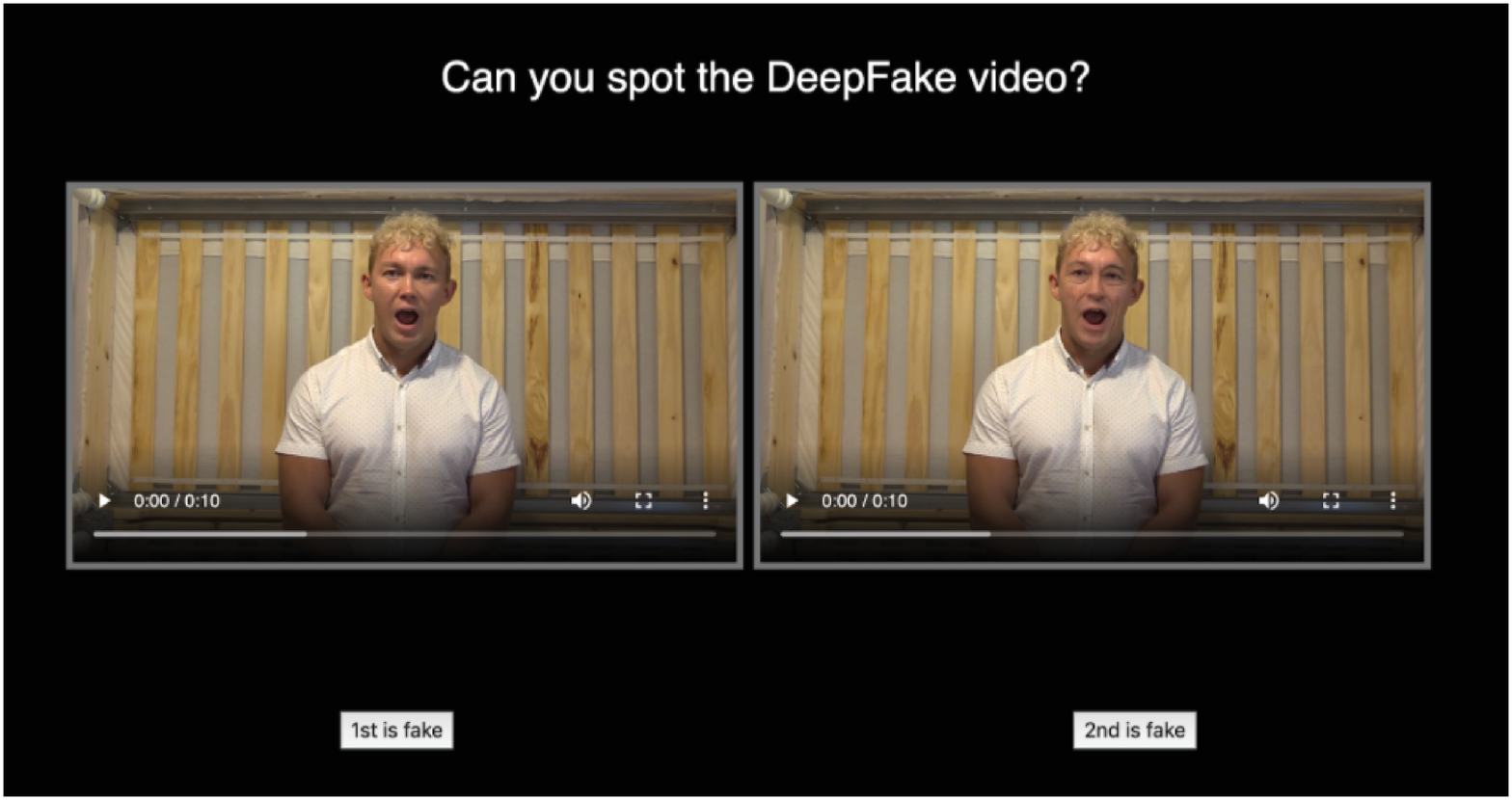}
    \includegraphics[width=0.85\textwidth]{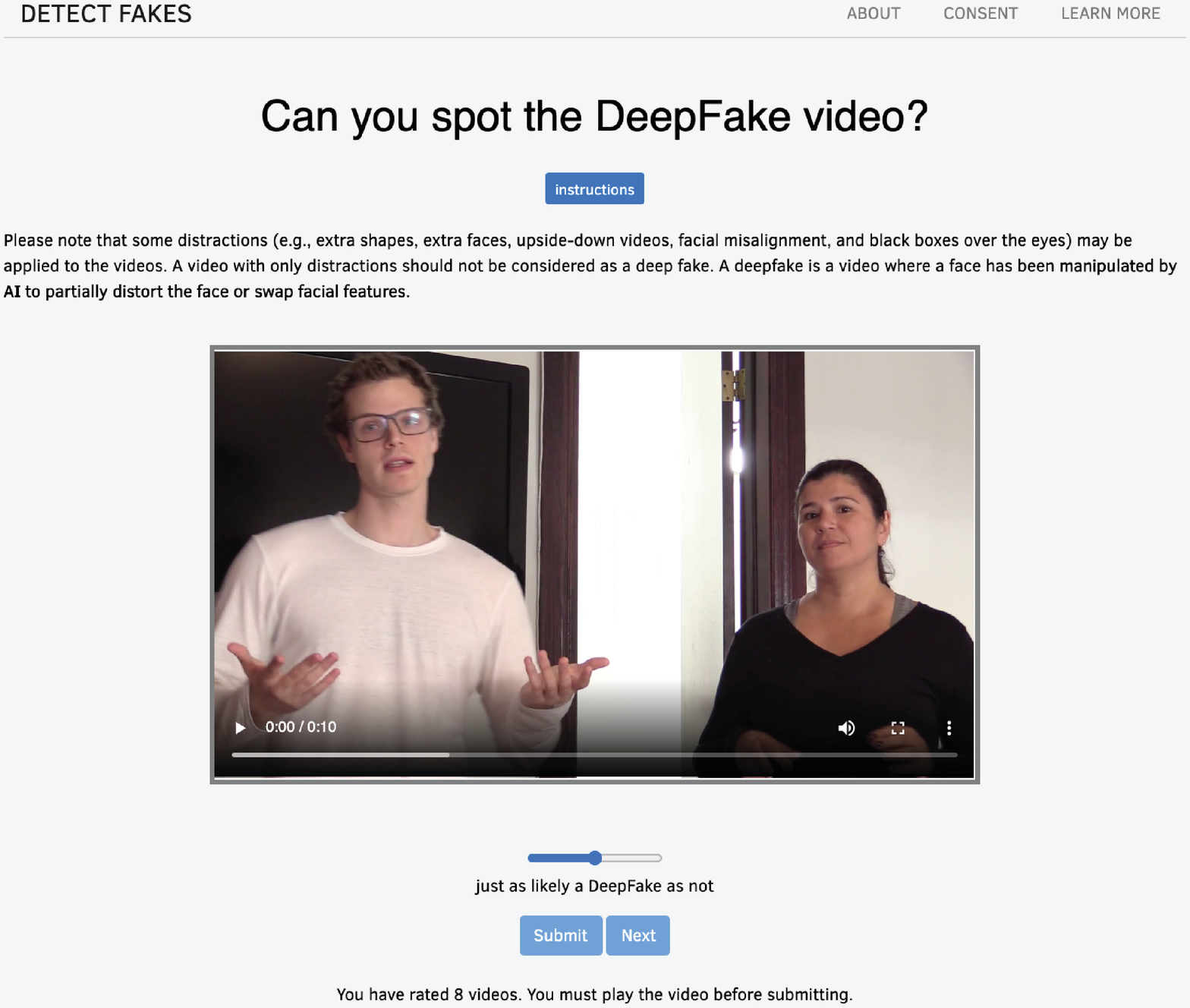}
    \caption{User interfaces for deepfake detection in Experiment 1 (two-alternative forced choice design) and Experiment 2 (single video design). In Experiment 2, we re-designed the color scheme of the website.}
    \label{fig:ux}
\end{figure*}

\end{document}